\pdfoutput=1

\documentclass[11pt]{article}

\usepackage[final]{acl}

\usepackage{times}
\usepackage{latexsym}
\usepackage{blindtext}
\usepackage{graphicx}
\usepackage{multirow} 
\usepackage{multicol} 
\usepackage{float}
\usepackage{booktabs}
\usepackage{svg}
\usepackage{pdfpages}

\usepackage{enumitem}
\usepackage{geometry}
\usepackage{lipsum}
\usepackage{graphicx}
\usepackage[T1]{fontenc}
\usepackage{subcaption}
\usepackage{comment}
\usepackage{url}

\usepackage[utf8]{inputenc}
\usepackage{subcaption}

\usepackage{microtype}

\usepackage{inconsolata}

\usepackage{graphicx}

\title{Open (Clinical) LLMs are Sensitive to Instruction Phrasings}

\author{Alberto Mario Ceballos Arroyo*$^\gamma$
\quad \textbf{Monica Munnangi*$^\gamma$} 
\quad \textbf{Jiuding Sun}$^\gamma$
\\
\quad \textbf{Karen Y.C. Zhang}$^\gamma$
\quad \textbf{Denis Jered McInerney}$^\gamma$$^\diamondsuit$ 
\quad  \textbf{Byron C. Wallace}$^{\gamma}$ 
\quad \textbf{Silvio Amir}$^\gamma$
  \\ 
$^\gamma$Northeastern University~~$^\diamondsuit$Codametrix\\
\texttt{\small \{ceballosarroyo.a, munnangi.m, 
sun.jiu, zhang.yuchen, b.wallace,s.amir\}@northeastern.edu} \\
\texttt{\small jmcinerney@codametrix.com}}

\begin{document}
\maketitle
\def\thefootnote{*}\footnotetext{Equal contribution}\def\thefootnote{\arabic{footnote}}

\begin{abstract}
Instruction-tuned Large Language Models (LLMs) can perform a wide range of tasks given natural language instructions to do so, but they are sensitive to how such instructions are phrased. 
This issue is especially concerning in healthcare, as clinicians are unlikely to be experienced prompt engineers and the potential consequences of inaccurate outputs are heightened in this domain. 

This raises a practical question: \emph{How robust are instruction-tuned LLMs to natural variations in the instructions provided for clinical NLP tasks?}
We collect prompts from medical doctors across a range of tasks and quantify the sensitivity of seven LLMs---some general, others specialized---to natural (i.e., non-adversarial) instruction phrasings. 
We find that performance varies substantially across all models, and that---perhaps surprisingly---domain-specific models explicitly trained on clinical data are especially brittle, compared to their general domain counterparts. 
Further, arbitrary phrasing differences can affect fairness, e.g., valid but distinct instructions for mortality prediction yield a range both in overall performance, and in terms of differences between demographic groups.

\end{abstract}
\section{Introduction}

Modern LLMs---e.g. \textsc{GPT-3.5+} \citep{radford2019language,ouyang2022training}, the FLAN series \citep{chung2022scaling}, Alpaca \citep{alpaca}, Mistral \citep{jiang2023mistral}---can execute arbitrary tasks \emph{zero-shot}, i.e., provided with only instructions  
rather than explicit training examples. 
LLMs have also shown promising improvements in performance on classification and information extraction (IE) tasks, such as named entity recognition \cite{brown2020language, munnangi2024onthefly} and relation extraction \citep{wadhwa2023revisiting, ashok2023promptner,jiang2024toner} in both general and specialized domains like biomedical and scientific literature \citep{agrawal2022large, wadhwa2023jointly,asada2024enhancing}. 

\begin{figure}[!t]
\centering
    \includegraphics[width=\columnwidth]{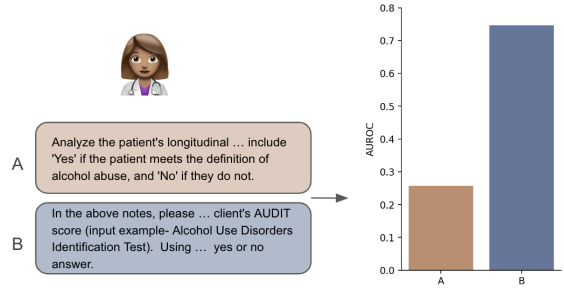}
  \vspace{-0.3cm}
  \caption{How much does LLM performance on clinical tasks depend on the arbitrary phrasings of instructions? Here we show an illustrative example: Discrepancy in AUROC score for \textsc{Clinical Camel} on the cohort selection-alcohol abuse classification task, when given the worst (A) and the best (B) performing prompts for \textsc{Alcohol-Abuse} classification task.}
  \label{fig:example}
\end{figure}

\begin{figure*}[!ht]
    \centering
    {\includegraphics[width=0.75\textwidth]{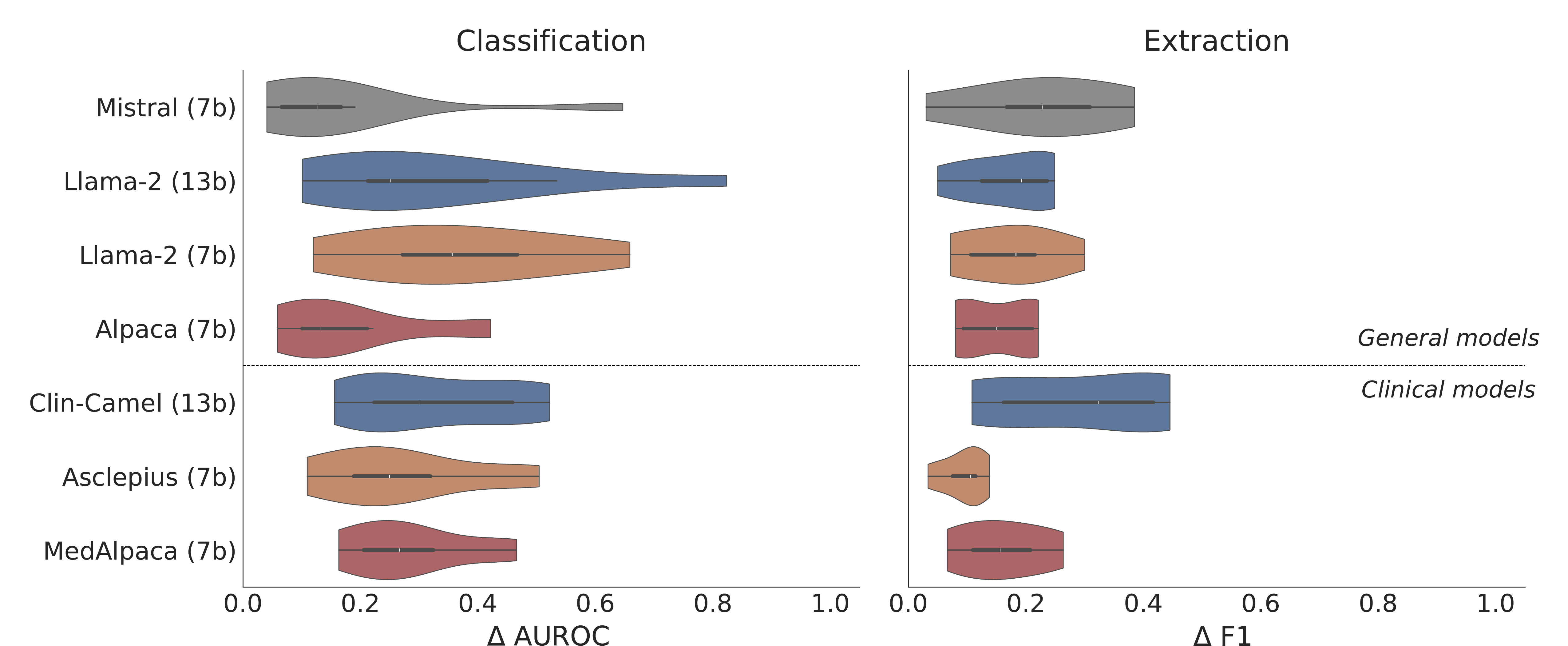}}
    \vspace{-0.3cm}
    \caption{Variance in performance for clinical classification and information extraction tasks for each model. We show the distribution of \textbf{deltas between the best and worst performing prompt} for each task.} 
    \label{fig:deltas}%
\end{figure*}

However, prior work has shown that LLMs do not ``understand’’ prompts~\citep{webson_prompt-based_2022} and are sensitive to the particular phrasings of instructions \citep{lu_fantastically_2022,sun_evaluating_2023}. 
Domain experts in specialized domains such as medicine are especially likely to interact with models by providing instructions (i.e., in \emph{zero-shot} settings), and are unlikely to be talented prompt engineers. 
For instance, a clinician might task a model to ``Extract and summarize the findings of the patient's last X-ray'', or ask ``When did the patient last receive a painkiller?''.  
It is unrealistic to fine-tune models for every possible such task; hence the appeal of models responsive to arbitrary prompts.
A downside, however, is that a clinician's particular phrasing may dramatically affect model performance (Figure \ref{fig:example}).
Such unpredictability is especially troublesome in healthcare, where poor performance might ultimately impact patient health.

In this work we ask: {\bf How sensitive are LLMs---general and domain-specific---to plausible instruction phrasing variations for clinical tasks?} 
Our analysis deepens prior work on robustness by focusing on the clinical domain;  
this is important both due to the higher stakes and because clinical notes differ qualitatively from general domain text. 
For example, notes in EHR often contain grammatical errors (``\textit{Pt complains of headache, and feel dizzy.}''); abbreviations not defined in context (``\textit{Pt}'' could be ``\textit{patient}'' or ``\textit{Prothrombin time}''), and; domain-specific jargon (``\textit{edema}'', ``\textit{Diuretic}'').

Therefore, one of the key aspects we consider is the domain-specificity of models. 
Are clinical LLMs more (or less) robust to different valid instruction phrasings written by doctors, compared to their general domain counterparts? 
To assess this, we evaluate recently released LLM variants trained on synthetic datasets comprising automatically generated clinical notes~\citep{kweon2023publicly}, and medical dialogue from case reports found in biomedical literature ~\citep{toma2023clinical}. 
We find that performance varies substantially given alternative instruction phrasings for both general and clinical LLMs. 
Figure \ref{fig:deltas} shows the distribution of deltas between the best and worst performing prompts across a set of clinical classification and information extraction tasks. 

Finally, we investigate how instruction phrasings impact the fairness of predictions, by which here we mean observed differences in performance between demographic subgroups. The degree to which LLMs might perpetuate and exaggerate such disparities in clinical use is a topic of active research \citep{omiye2023large,pal2023bias,zack2024assessing}.
Here we contribute to this by investigating the interaction between prompt phrasings and fairness. 
We find significant performance differences (up to 0.35 absolute difference in AUROC) in a mortality prediction task from MIMIC-III between \textit{White} and \textit{Non-White} subgroups and also a significant disparity between \textit{Male} and \textit{Female} patients (up to 0.19 absolute difference in AUROC). To facilitate future research in this direction, we release our code and prompts\footnote{\url{https://github.com/alceballosa/clin-robust}}.

\section{Experimental Framework}

Our experimental setup is intended to quantify the robustness of LLMs to natural variations in instructional phrasings for clinical tasks. 
We considered a set of ten clinical classification tasks and six information extraction tasks drawn from MIMIC-III~\cite{Johnson2016MIMICIIIAF} and prior i2b2 and n2c2 challenges,\footnote{\url{https://n2c2.dbmi.hms.harvard.edu/}} summarized in Table \ref{tab:dataset} (\S \ref{sec:tasksdata}).  
We recruited a diverse group of medical professionals to write prompts for each task (\S \ref{sec:instructions}). 
We then evaluated the performance, variance, and fairness of seven LLMs (four general-domain and three domain-specific) across prompts (\S \ref{sec:models}). 

\subsection{Tasks and Datasets}

\label{sec:tasksdata}
\begin{table*}[h!]
    \centering
    \small
    \begin{tabular}{l c c c c}
    \toprule
     \textbf{Dataset} & \textsc{Task} & \textsc{Test Set} & \textsc{Task type}\\

         \midrule
         \multirow{1}{*}{MIMIC-III} & In-hospital Mortality & 160 & 
            Binary Classification \\
         \cmidrule{2-5}
         \multirow{4}{*}{Obesity co-morbidity} & Asthma & 507 & Binary Classification  \\
         & CAD & 507 & Binary Classification \\
         & Diabetes & 507 & Binary Classification \\
         & Obesity & 507 &  Binary Classification \\
         \cmidrule{2-5}
         \multirow{3}{*}{Cohort Selection} & Abdominal & 86 & Binary Classification &\\
         & Alcohol-Abuse & 86 & Binary Classification \\
         & Drug-Abuse & 86 & Binary Classification \\
         & English & 86 & Binary Classification  \\
         & Decisions & 86 & Binary Classification \\
         \cmidrule{2-5}
         \multirow{1}{*}{Medical Challenge}  & Medication & 251 & Extraction\\
         \cmidrule{2-5}
         \multirow{4}{*}{Relation Challenge}  & Concept Problem & 256 & Extraction\\
         & Concept Test & 256 & Extraction \\
         & Concept Treatment & 256 & Extraction \\
         \cmidrule{2-5}
         \multirow{1}{*}{Adverse Drug Effects} & Drug & 202 & Extraction \\
         \cmidrule{2-5}
         \multirow{1}{*}{Risk Assessment} & Risk Factor CAD & 514 & Extraction \\

         \bottomrule
    \end{tabular}
    \caption{Tasks and datasets used for evaluation.}
    \label{tab:dataset}
\end{table*}

\paragraph{MIMIC-III \citep{Johnson2016MIMICIIIAF}} is a database of de-identified EHR comprising over 40k patients admitted to the intensive care unit of the Beth Israel Deaconess Medical Center between 2001 and 2012.
It comprises structured variables and clinical notes (e.g., doctor
and nursing notes, radiology reports, discharge
summaries); we focus on the latter. 
MIMIC-III also contains demographic information, including ethnicity/race, sex, spoken language, religion, and insurance status \citep{Chen2019CanAH}. 
As an illustrative predictive task, we consider in-hospital mortality prediction, which has been the subject of prior work \citep{Harutyunyan2017MultitaskLA}. 
Owing to compute constraints, we sub-sampled the test-split to 10\% of the data (preserving class ratio), yielding 160 records for evaluation. 



\paragraph{n2c2 2018 Cohort Selection Challenge \citep{1496751}} aims to identify whether a patient meets the criteria for inclusion in a clinical trial based on their longitudinal records. The dataset contains 288 patients, their associated clinical notes 
and a set of binary labels indicating whether they meet the criteria for each of 13 possible cohorts (e.g., drug abuse, alcohol abuse, ability to make decisions, among others). In this study, we focus on the 5 cohorts shown in Table \ref{tab:dataset} and treat each as an independent binary classification task aiming to predict whether the criteria is ``met'' or ``not met''.

\paragraph{i2b2 2008 Obesity Challenge \citep{1498659}} entails identifying patients suffering from obesity and its co-morbidities from
their discharge summary notes. 
The dataset comprises 1027 pairs of de-identified discharge summaries and 16 disease labels 
from intuitive judgements which are based on the entire discharge summary.
We report the performance for obesity and three co-morbidities (i.e., asthma, atherosclerotic cardiovascular disease (CAD), and diabetes mellitus (DM)), each framed as a binary classification task aiming to predict whether the condition is ``present'' or ``absent''.

\paragraph{n2c2 2018 Adverse Drug Events and Medication Extraction in EHRs \citep{henry2018N2c2Shared2020a}} consists of a relation extraction task focused on identifying drugs/medications and their relations to adverse events for the patient. The dataset contains 202 patients and we focus only on the named entity recognition portion of the task (i.e. recognizing spans referring to drugs/medications).

\paragraph{i2b2 2014 Identifying Risk Factors for Heart Disease over Time \citep{stubbsIdentifyingRiskFactors2015}:} entails identifying medical risk factors linked to Coronary Artery Disease (CAD) in the EHR of patients with diabetes. The target factors include hypertension, obesity, smoking status, diabetes, hyperlipidemia, family history, and CAD itself. Here we consider only the latter.

\paragraph{i2b2 2010 Relations Challenge \citep{uzuner2010I2b2VA2011a}} consists of three related tasks: (1) identification of medical problems, tests, and treatments; (2) classification of assertions made on medical problems; and (3) relation extraction concerning medical problems, tests, and treatments. 
The data for this challenge includes discharge summaries from Partners HealthCare, and the Beth Israel Deaconess Medical Center \citep{6092050}, as well as discharge summaries and progress notes from the University of Pittsburgh Medical Center. We conduct evaluation on the first task (i.e. extraction of problems, tests, and treatments) over the notes of 256 patients.

\paragraph{i2b2 2009 Medication Extraction Challenge \citep{patrickHighAccuracyInformation2010}} focuses on the extraction of medications from clinical notes in the EHR, as well as their modes, reasons and frequency of administration. We center our analysis on medication extraction only, which encompasses around 1250 unique medications over 251 notes.

\subsection{Instruction Collection}
\label{sec:instructions}

We hired twenty medical professionals from different professional and demographic backgrounds, with varying medical specialties and years of experience. 
These included medical doctors (physicians, surgeons), medical writers/editors, nurses, and medical consultants from various countries, such as the United States, Nigeria, Kenya, Canada, Zambia, Egypt, Malawi, Pakistan, Philippines, and Ethiopia. All participants were either native-speakers or proficient in English. It should also be noted that participants were not required to have experience with LLMs but the majority of them reported having used these models in the past.

We provided participants with a description of the tasks including the goal, the expected outputs and a (fictitious) example of a clinical note. 
We then asked them to write instructions (in English) for each task with the only constraint being that they had to ensure the model outputs a valid label (for classification tasks) or a list of items (for extraction tasks). 
Figure \ref{fig:anno_ex} (Appendix \ref{apx:anno}) shows an example of the instructions given for a classification task. 

Initially, we ran a smaller scale pilot study consisting of one classification and one extraction task, and recruited participants who successfully completed the tasks. 
The process took around 5 hours on average and we compensated each participant at a rate of \$25/hour. 
We manually reviewed all written instructions and found that some were of poor quality (e.g., did not adhere to the goals of the task, or did not ensure that the model outputs valid responses). 
In such cases, we removed the author from the study and discarded all of their instructions. 
We also removed everyone that did not complete all the tasks, resulting in a final collection of instructions from 12 participants. See Appendix~\ref{apx:anno} for illustrative examples of the collected instructions\footnote{The full set of instructions is available in our code repository}.

\begin{figure*}[!ht]
\includegraphics[width=0.5\linewidth]{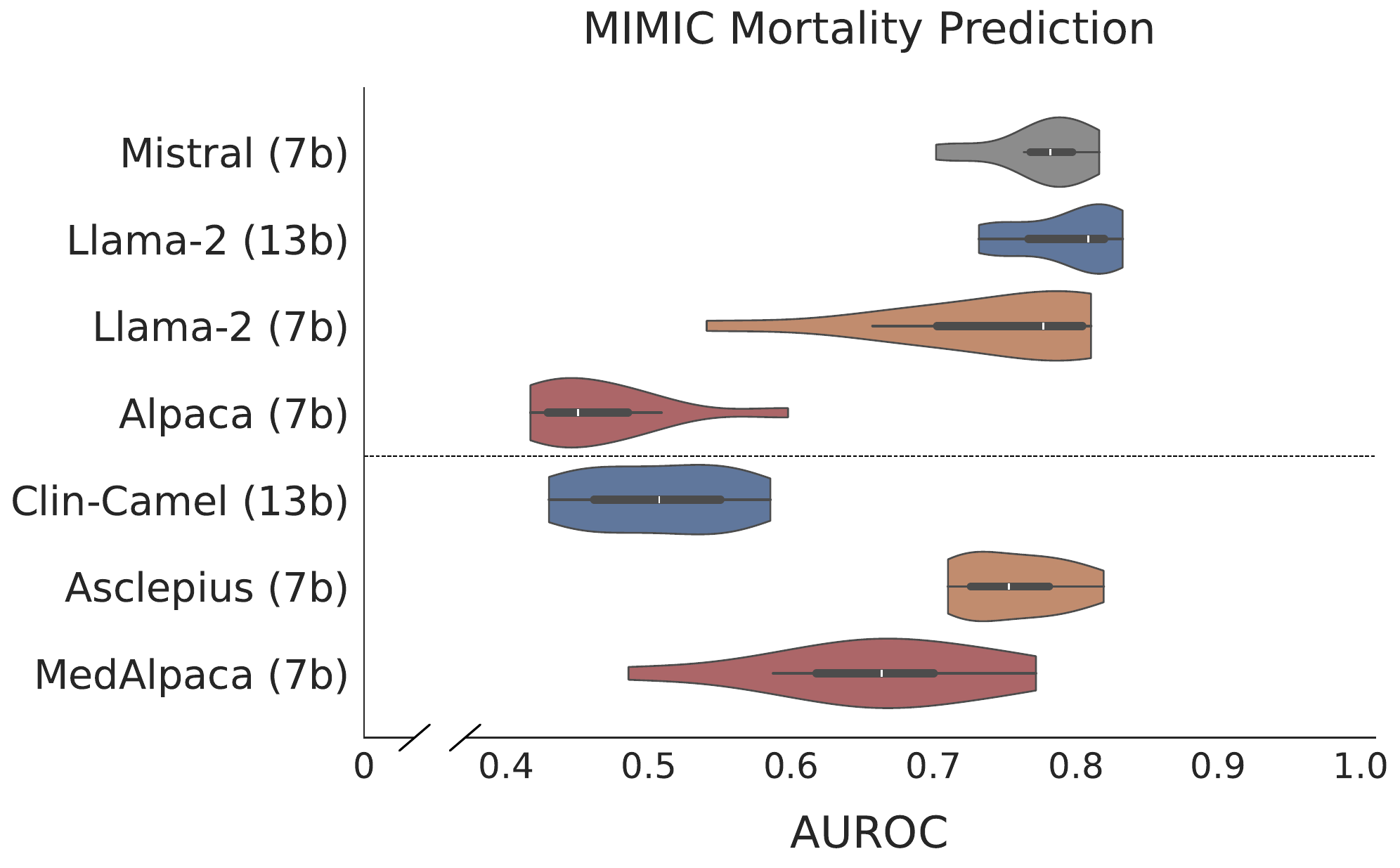} 
  \includegraphics[width=0.3553\linewidth]{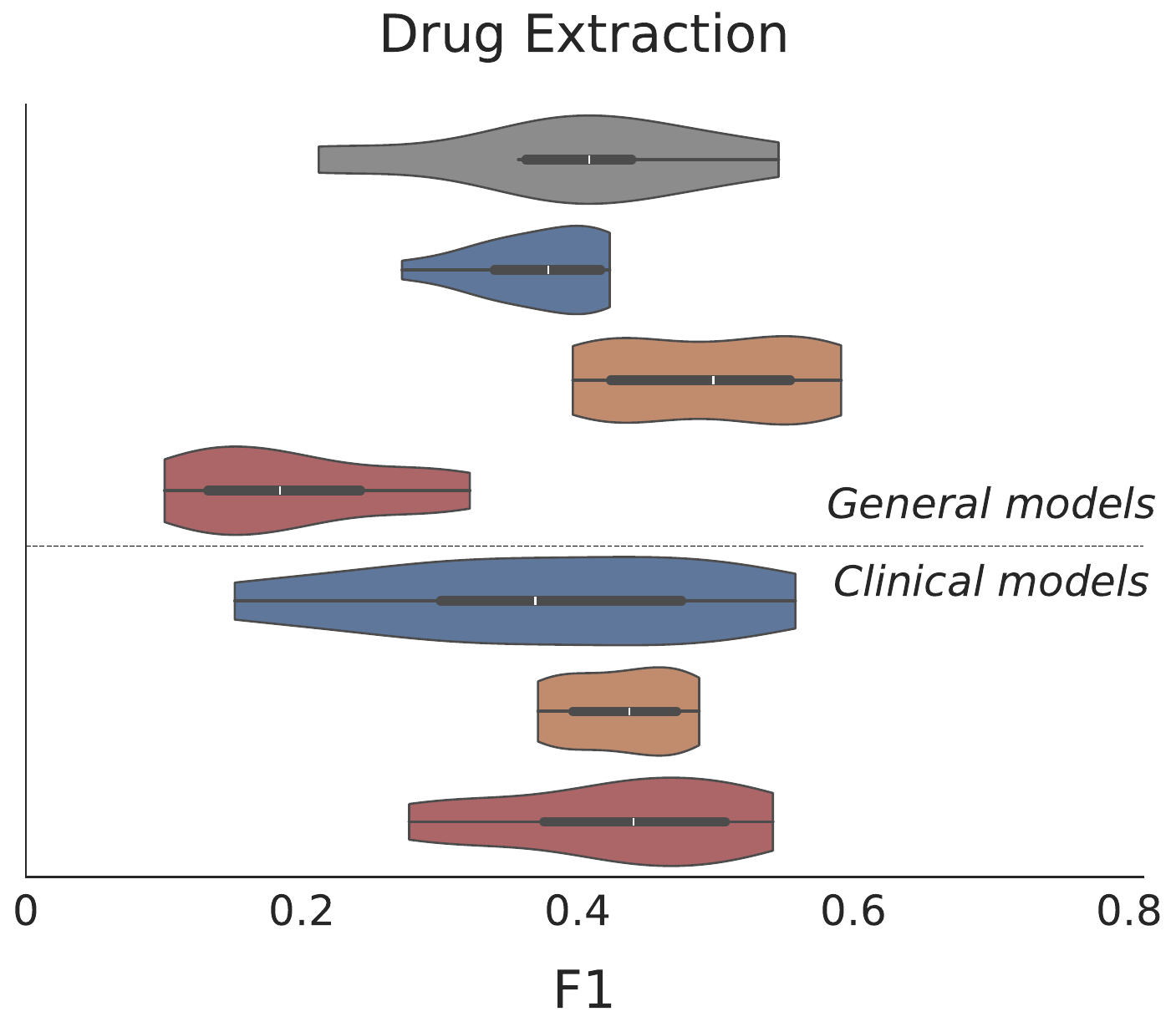}
  \vspace{-0.3cm}
  \caption {Variability in performance across prompts for the mortality prediction and drug extraction tasks. For most models, different but semantically equivalent prompts yield quite a range of performance.}
  \label{fig:cls_ext} 
\end{figure*}

\subsection{Models}
\label{sec:models}

We measured the performance, variance and fairness of seven general and domain-specific LLMs on each task, using the instructions written by medical professionals. To assess the impact of clinical instruction tuning, we paired all clinical models with their general domain counterparts. We considered three clinical models: \textsc{Asclepius (7b)}~\citep{kweon2023publicly}, \textsc{Clinical Camel (13b)}~\citep{toma2023clinical}, and \textsc{MedAlpaca (7b)}~\citep{han2023medalpaca}; and their corresponding base models, i.e., \textsc{Llama 2 Chat (7b)}, \textsc{Llama 2 Chat (13b)}~\citep{touvron2023llama}, and \textsc{Alpaca (7b)}~\citep{alpaca}, respectively.
We also included \textsc{Mistral IT 0.2 (7b)}~\citep{jiang2023mistral} in our experiments due to its high performance in standard benchmarks.


For all models and datasets, we performed zero-shot inference via prompts with a maximum sequence length of 2048 tokens which included the instruction, the input note, and the output tokens (64 for classification, 256 for extraction).
Since most clinical notes were too long to process in a single pass, we followed \citealt{huang2020clinicalbert} and split each note into chunks to be processed independently.  
For binary classification and prediction tasks, we treated the output for a given input note as positive if at least one of the chunks was predicted to be positive, and negative otherwise. For extraction tasks, we combined the outputs from each chunk into a single set of extractions.  

\paragraph{Evaluation:} Evaluation with generative models is challenging: Models may not respect the desired output format, or may generate responses that are semantically equivalent but lexically different from references~\citep{wadhwa2023jointly,agrawal2022large}.
We therefore took predictions from the output distribution of the first generated token by selecting
the largest magnitude logit from the set of target class tokens. For extraction tasks, we parsed generated outputs and performed exact match comparison with target spans.
We report ${\textrm{AUROC}}$ scores for classification tasks and F1 scores for extraction tasks.

\section{Results}


We present our main results for Mortality Prediction and Drug Extraction in Figure \ref{fig:cls_ext} --- results for the other classification and information extraction tasks can be found in Appendix \ref{apx:results}, Figures \ref{fig:classification} and \ref{fig:extraction}, respectively. Most models
show significant variability in performance for alternative but semantically equivalent instructions in both classification and extraction tasks. 
To further examine these observed disparities, we plotted the distribution of deltas between the best and worst performing prompts for each task in Figure \ref{fig:deltas}. 
We see that performance deltas can go up to 0.6 absolute ${\textrm{AUROC}}$ points for classification tasks and up to 0.4 absolute F1 points for extraction tasks.

In the Mortality Prediction task, we find that \textsc{Llama 2 (13b)} outperforms all other models, including the domain-specific ones (Figure \ref{fig:cls_ext}). 
However, for the other classification tasks, \textsc{Mistral} yields the best results often outperforming the larger models whilst exhibiting less variance (Figure \ref{fig:classification}). 
Regarding the clinical models, we observe that \textsc{Asclepius} consistently attains the best performance in classification tasks albeit with comparable variance. 

In the Drug Extraction task, \textsc{Llama 2 (7b)} attains the best results on average but with comparable variance to other general LLMs. 
However, the results for clinical models are mixed: while \textsc{Clinical Camel} can achieve the highest performance given the best prompt, it also has the highest variance and lowest median performance. 
\textsc{MedAlpaca} comes close to \textsc{Clinical Camel} in the best case scenario but with less variance and better median performance. 
\textsc{Asclepius} has a median performance similar to that of \textsc{MedAlpaca} but with a much lower variance. 
We observe similar trends for the other information extraction tasks: \textsc{Llama 2 (7b)} consistently outperforms other general LLMs with similar variance, whereas none of the clinical models is clearly superior across tasks --- however, \textsc{Asclepius} seems to have the least variance overall.

\begin{figure*}[h]
  \centering 
  {\includegraphics[width=0.8\textwidth]{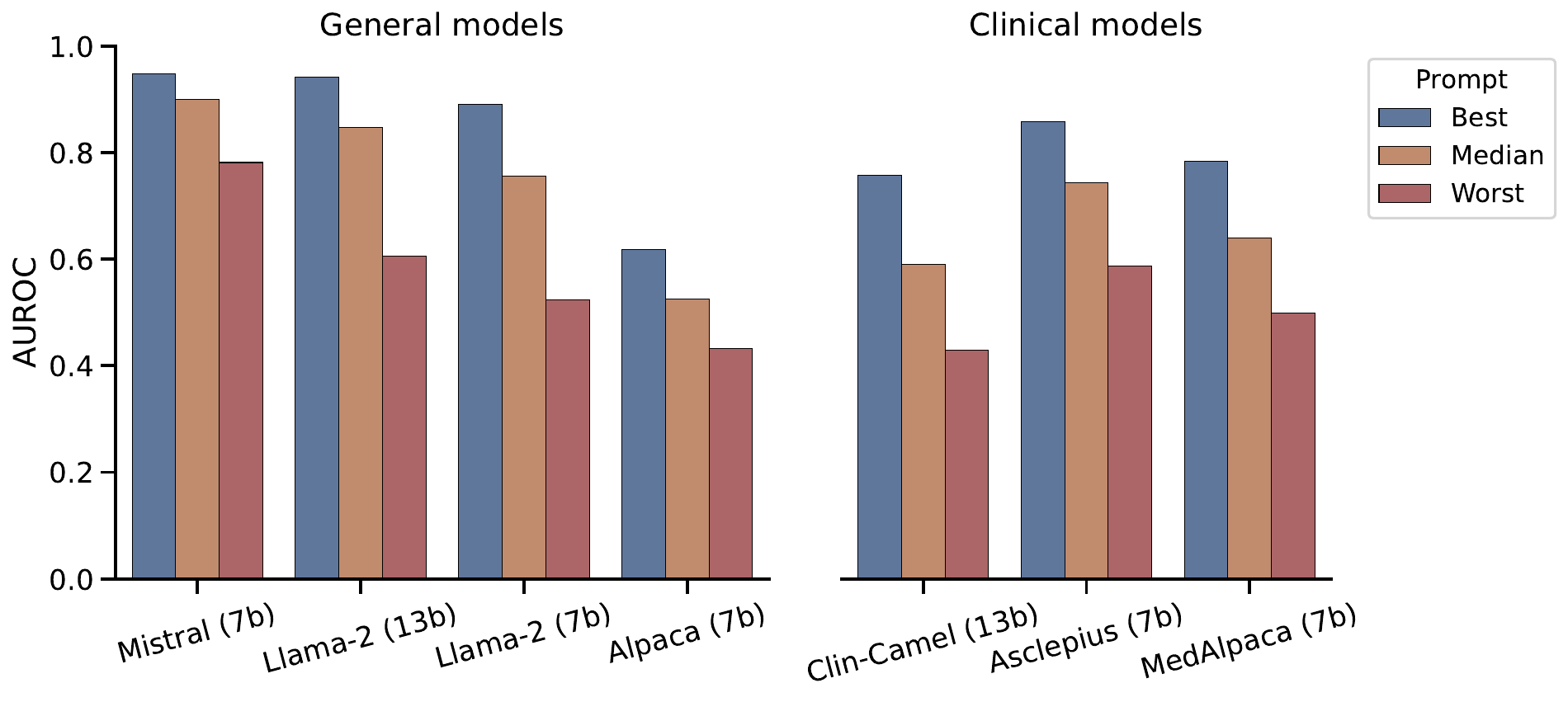}}
 \vspace{-0.3cm}
  \caption{Average AUROC across classification tasks given the best, median, and worst-performing prompts for each model.}
  \label{fig:rankingc} 
\end{figure*}

\begin{figure*}[h]
  \centering 
    {\includegraphics[width=0.8\textwidth]{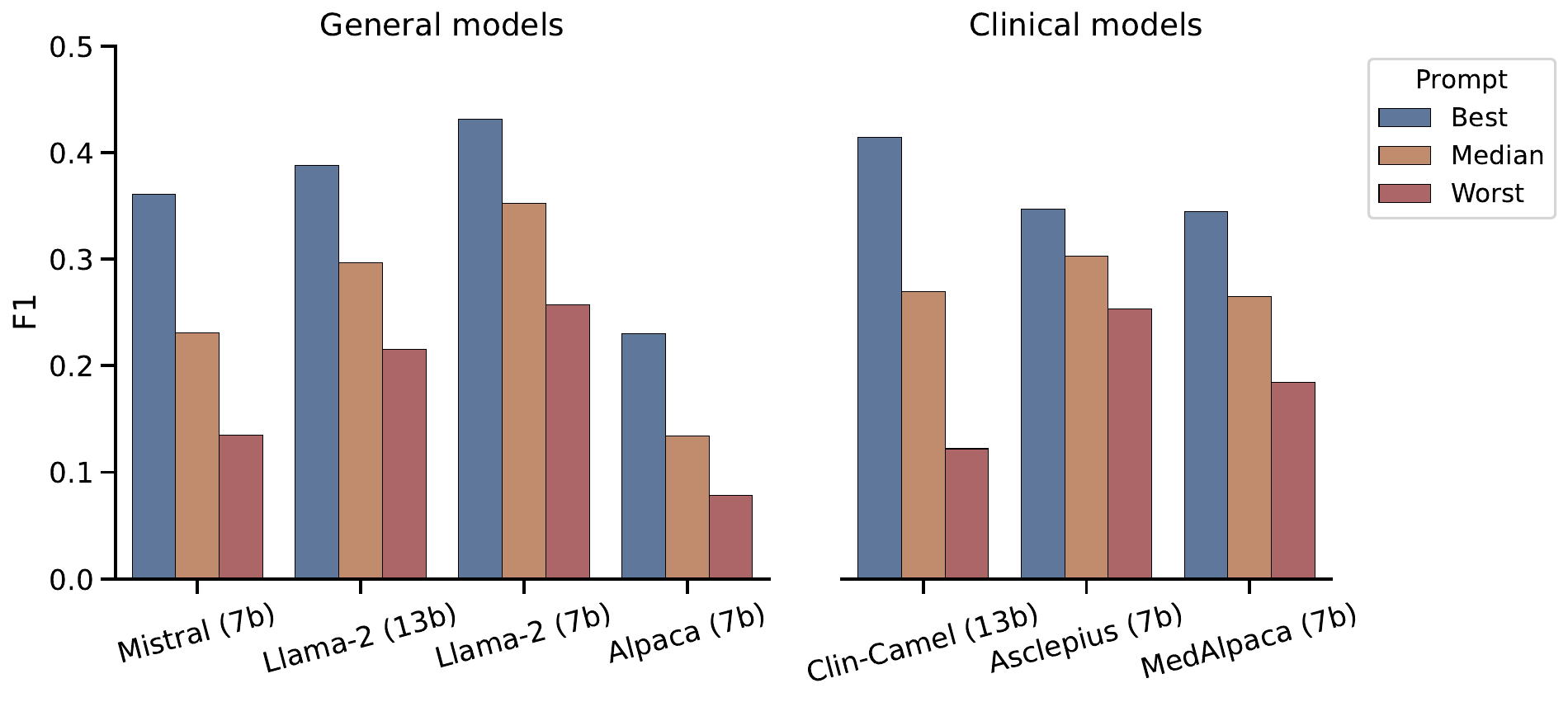}}
    \vspace{-0.3cm}
  \caption{Average F1 across extraction tasks given the best, median, and worst-performing prompts for each model.}
  \label{fig:rankinge} 
\end{figure*}

To better understand the differences between the general domain and clinical LLMs, we compared their average performance given the best, median and worst prompts. 
Figures \ref{fig:rankingc} and \ref{fig:rankinge} show the results per model averaged across all classification and extraction tasks, respectively. 
Surprisingly, we find that general domain models outperform their domain-specific counterparts --- with the exception of \textsc{Alpaca} which performs poorly across all tasks.
Again we observe that even though \textsc{Clinical Camel} \emph{can} outperform its general domain analog in extraction tasks given the best prompt, it also shows more variance and much lower performance in the worst case.

\begin{figure*}[h]
  \centering 
    {\includegraphics[width=0.8\textwidth]{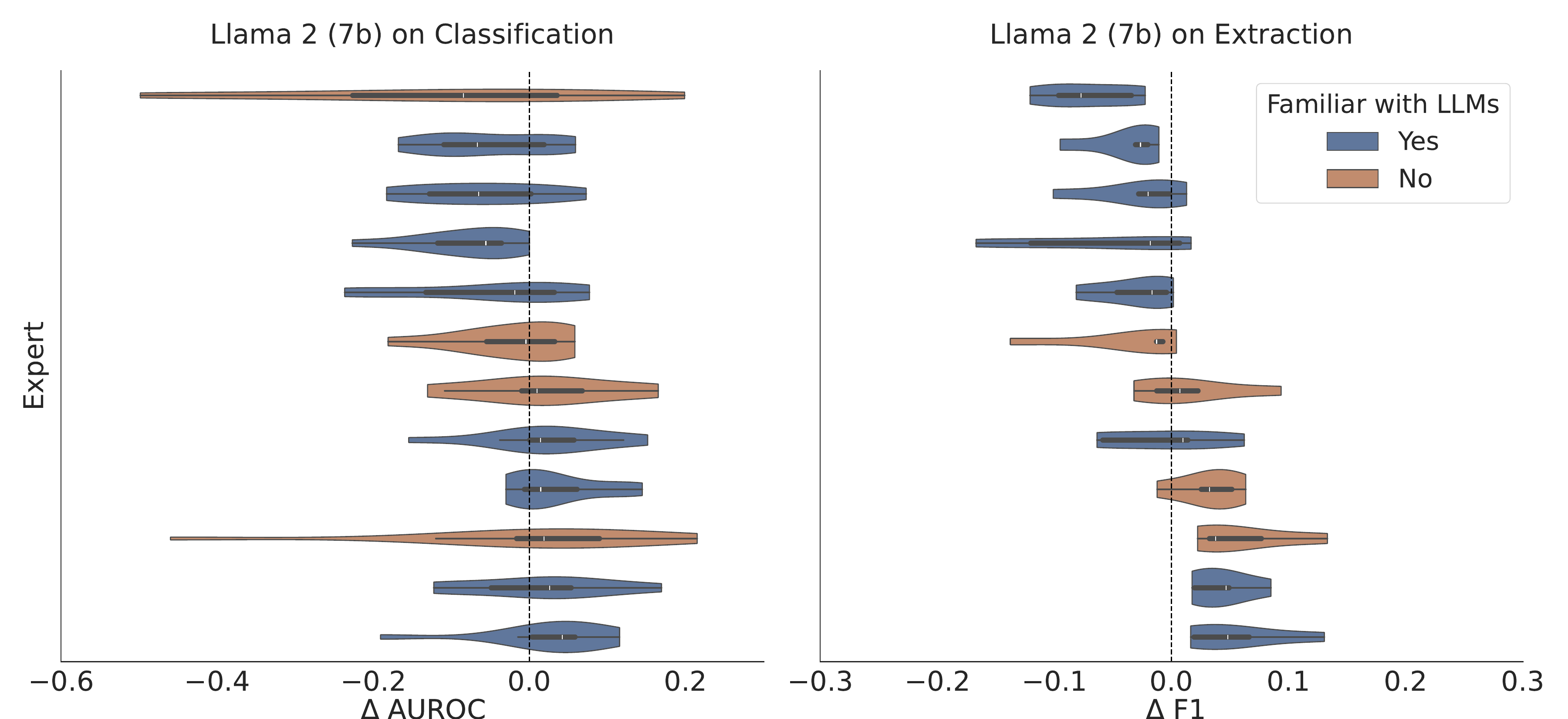}}
  \caption{Distribution of performance deltas between each expert’s prompt and the median prompt across all tasks. Each violin plot represents an expert color coded according to their familiarity with LLMs. }
  \label{fig:annotator_deltas} 
\end{figure*}

Finally, we investigated whether the observed performance variability can be explained by individual differences between experts in prior experience with LLMs or aptitude in writing effective instructions. To assess this, we measured the performance deltas between each prompt and the median prompt for each classification and extraction task. Figure \ref{fig:annotator_deltas} shows the results for \textsc{Llama 2 (7b)} and results for other models can be found in Appendix \ref{apx:results}, figures \ref{fig:annotators_analysis_general} and \ref{fig:annotators_analysis_clinical}. We find that there are indeed significant differences at the individual level, both in terms of variance and overall performance, particularly for classification tasks. Only roughly half the users can (somewhat) consistently beat the median performance across tasks. We also note these differences can not be solely explained by prior experience with LLMs --- some novice users are able to consistently write more effective instructions as compared to other experienced users. However, one caveat is that this prior experience is most likely with larger commercial models which may be more robust to instruction variations.


\subsection{Fairness}
\label{sec:fairness}
\begin{figure}[th]%
    \centering
    \subfloat{{\includegraphics[width=7cm]{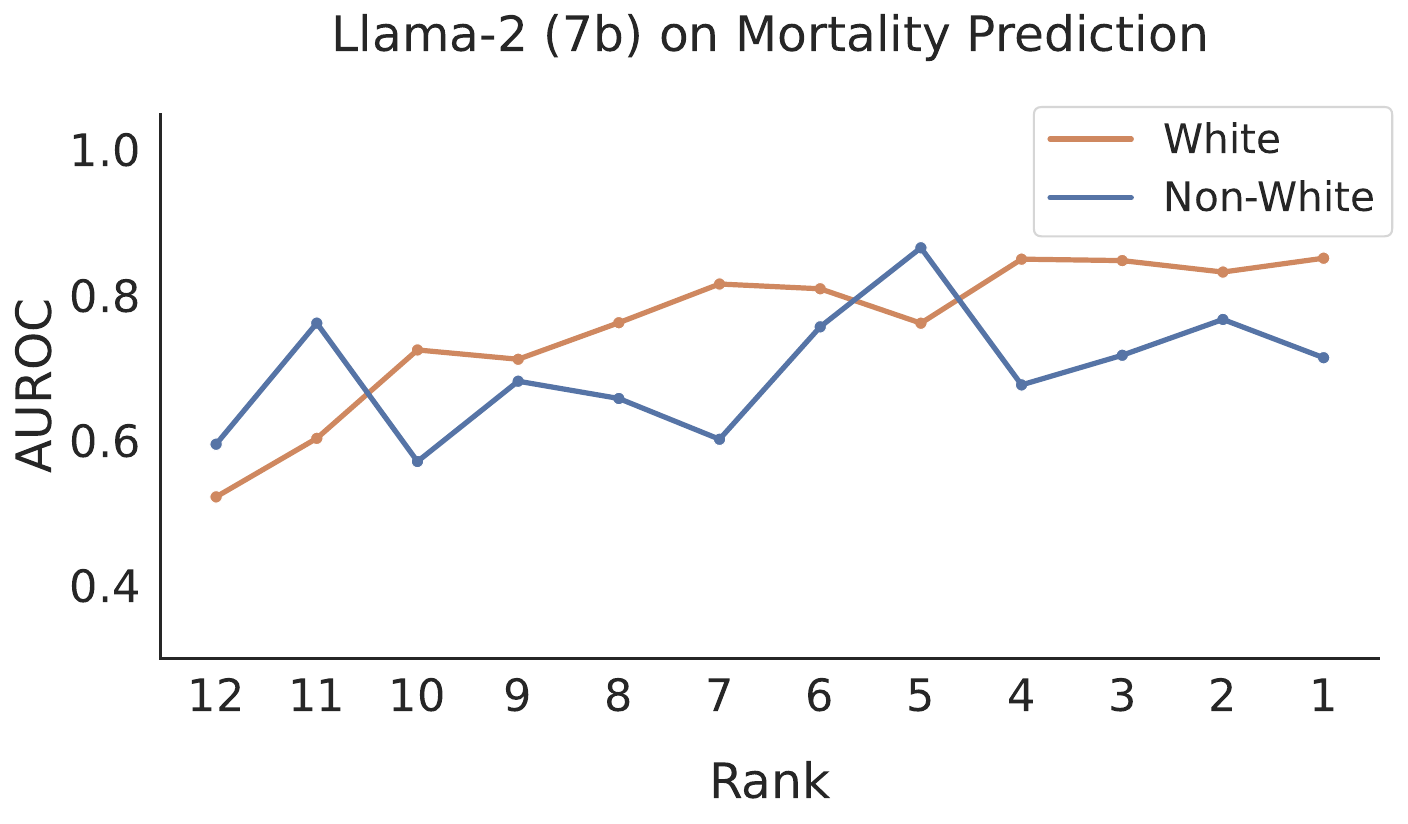} }}%
    \hspace{0em}
    \subfloat{{\includegraphics[width=7cm]{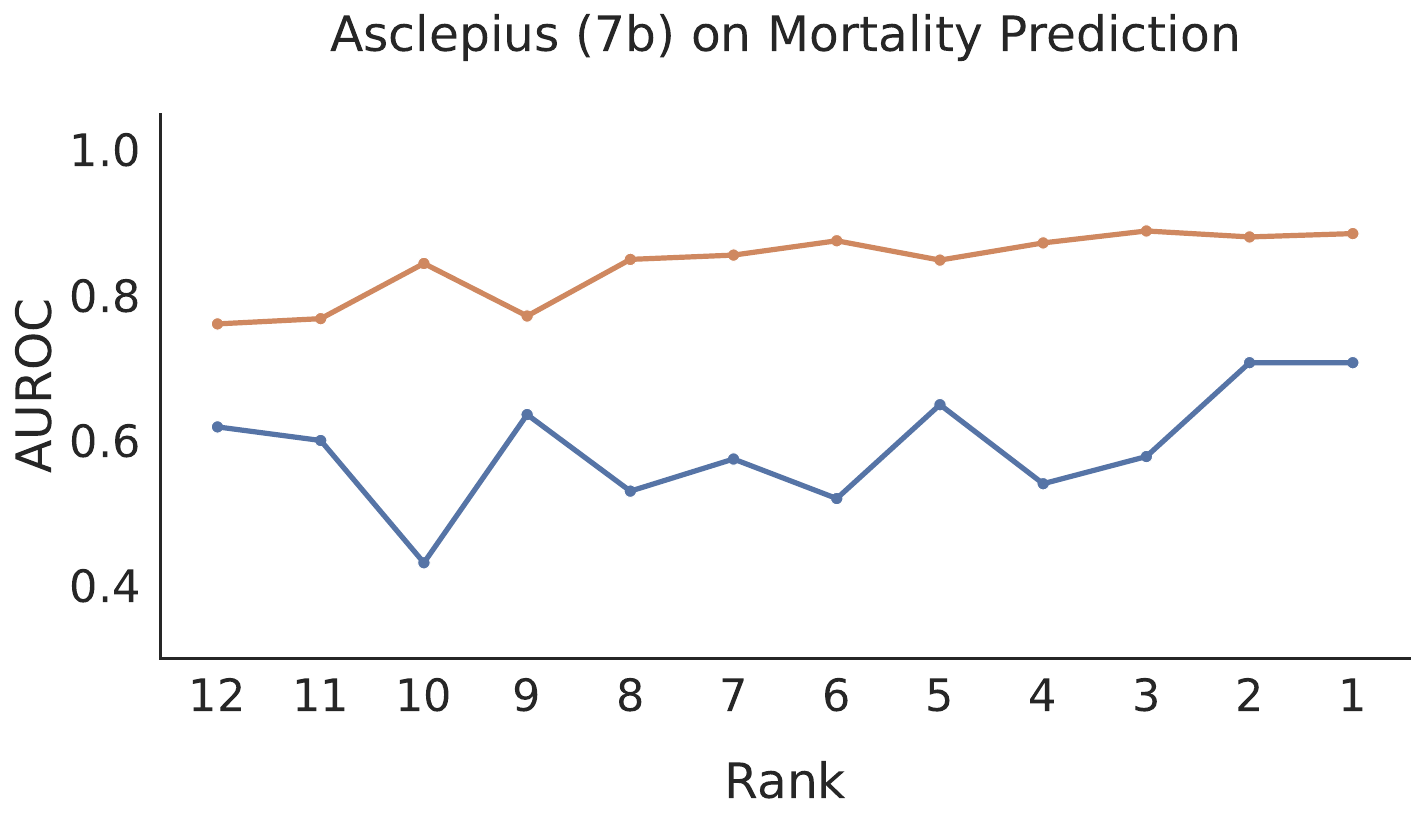} }}%
    \vspace{-0.2cm}
    \caption{Race subgroup performance on the Mortality Prediction task with a general (top) and clinical model (bottom)}%
    \label{fig:race_analysis}%
\end{figure}

\begin{figure}[t]%
    \centering
    \subfloat{{\includegraphics[width=7cm]{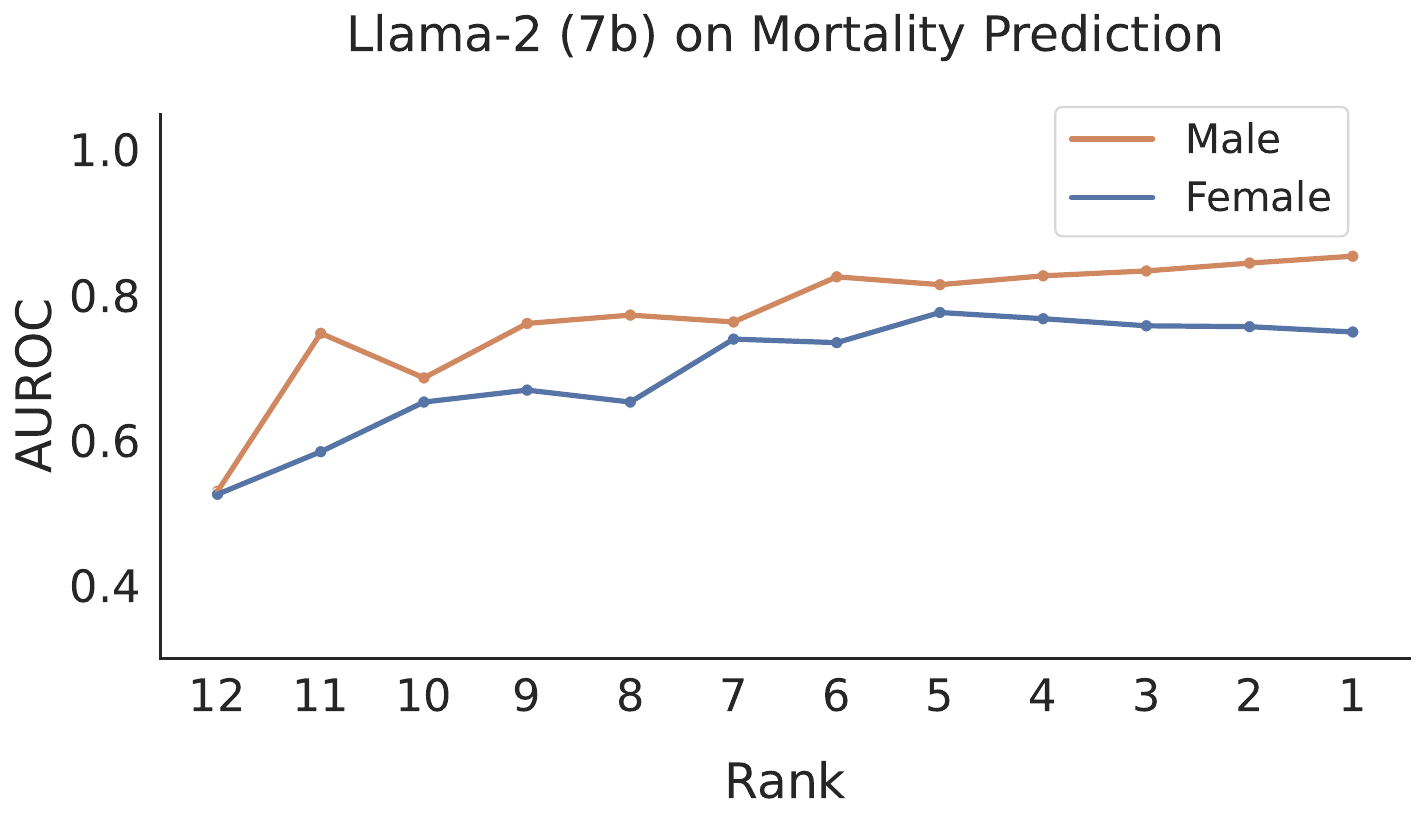} }}%
    \hspace{0em}
    \subfloat{{\includegraphics[width=7cm]{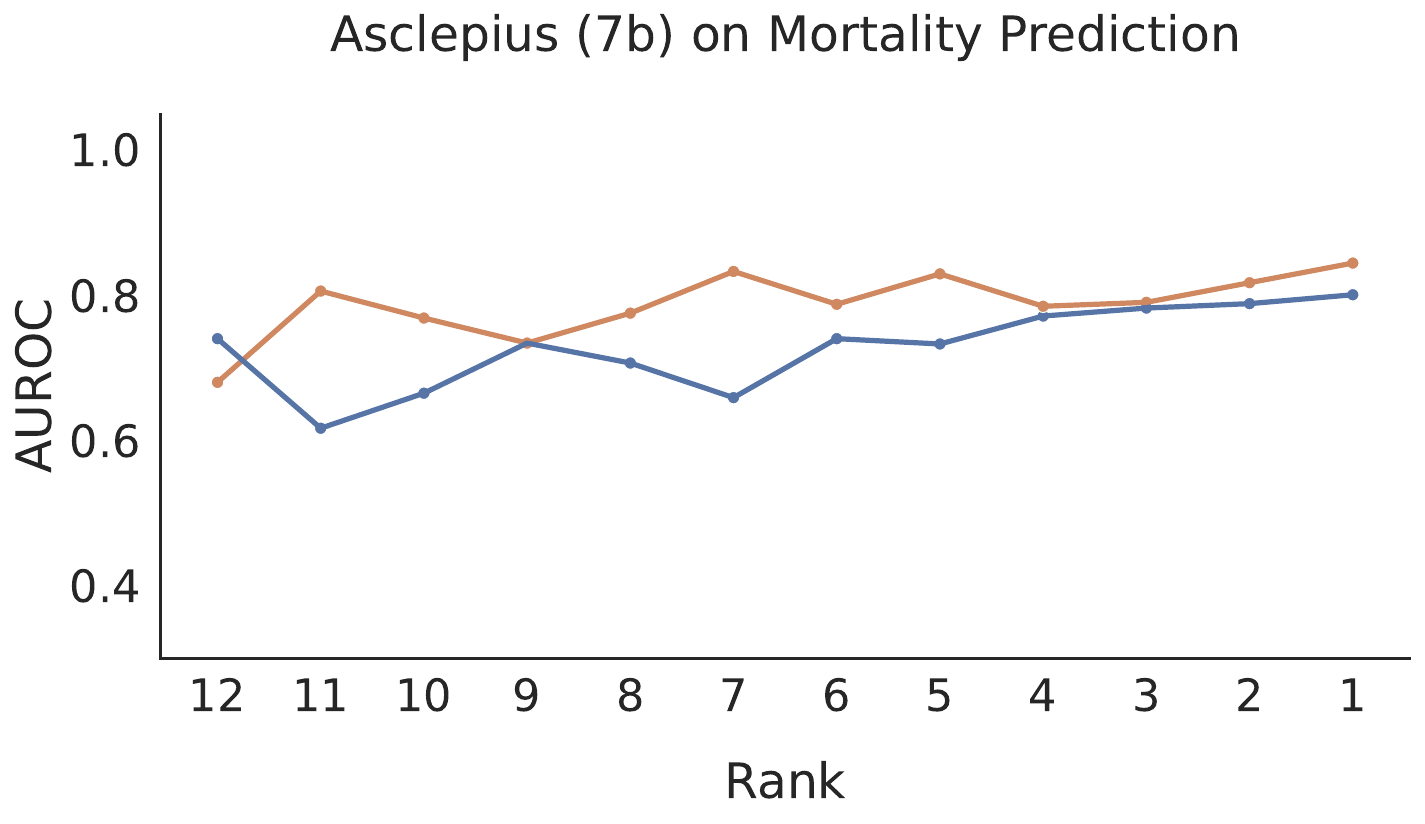}}}%
    \vspace{-0.2cm}
    \caption{Gender subgroup performance on the Mortality Prediction task with a general (top) and clinical model (bottom)}%
    \label{fig:sex_analysis}%
\end{figure}

How do variations in prompt phrasings impact model fairness (here measured as disparities in predictive performance for specific demographic subgroups)? 
To answer this question, we stratified the patients in the mortality prediction task with respect to race and sex. 
To avoid issues with reliability of performance metrics arising from small sub-samples~\citep{amir2021impact} we only consider two broad groups (i.e., \emph{White} and \emph{Non-White}). 
We sorted the instructions according to their overall performance and plot individual subgroup performance (Figure \ref{fig:race_analysis}). 
We repeated the analysis for sex (as indicated in EHR) and present individual subgroup performance in Figure \ref{fig:sex_analysis}.

\begin{table}[!t]
    \centering
    \small
    \begin{tabular}{l c c c c c}
    \toprule
               &                  & \multicolumn{2}{c}{\textbf{Gender}}   & \textbf{Total}   \\
               &                  & Female      & Male      &         \\ 
\midrule
\textbf{Race} & White    &   52  &  59   &   111 \\
               & Non-White    &   24  &   25    & 49  \\
\midrule
\textbf{Total}&    & 76 & 84 &  160 \\
\bottomrule
    \end{tabular}
    \caption{Distribution of gender and race in the sample used examine model fairness (\S \ref{sec:fairness})}
    \label{tab:race_and_sex}
\end{table}

In line with prior work \cite{amir2021impact, Adam_2022},  we observe that models have disparate performance for different subgroups. 
Both \textsc{Llama 2 (7b)} and \textsc{Asclepius (7b)} tend to under-perform for non-White patients compared to White counterparts with absolute differences of up to $0.21$ and $0.35$ AUROC points, respectively.
A possible explanation is that the way in which medical staff write clinical notes differ for White vs Black patients~\citep{Adam_2022}. 
However, here non-Whites are an heterogeneous group so there may be other confounding factors. 

In regards to sex, we again observe noticeable (albeit smaller) differences in performance with \textsc{Llama 2 (7b)} performing worse for \textit{Female} patients across all the prompts with relative differences of up to $0.16$ absolute AUROC points, and \textsc{Asclepius (7b)} yielding differences of up to $0.19$ points.
Overall, these results indicate that natural variations in prompts may translate to wide differences in fairness. 
Troublingly, a clinician using such models would likely be unaware that apparently benign phrasing changes may disproportionately affect particular demographic groups.

\subsection{Discussion}

Our experiments show that instruction-tuned LLMs are not robust to plausible variations in instruction phrasings --- equivalent but distinct instructions result in significant differences in both task performance and fairness with respect to demographic subgroups. Moreover, we find that no single model yields optimal performance across tasks, e.g. Mistral 7b is the best model for classification but has middling performance in extraction tasks. We also find that general domain models tend to outperform clinical models --- although surprising, these findings corroborate prior work on clinical text summarization~\cite{Veen2023AdaptedLL}. This may be due to the fact that clinical models are fine-tuned with synthetic or proxy data that does not adequately capture the idiosyncrasies of clinical notes from EHR. 

\section{Related Work}


\paragraph{Instruction-following LLMs} 
Scaling up decoder-only language models imbues them with the ability to solve various tasks given only instructions or a small set of examples at inference time~\cite{brown2020language,chowdhery2022palm}. Follow-up work sought to improve this by explicitly training GPT-3 to follow instructions and provide helpful and harmless responses via Reinforcement Learning from Human Feedback~\cite{ouyang2022training,openai2022chatgpt}. Others showed that fine-tuning with a causal language modeling objective over labeled data formatted as instruction/response pairs is sufficient to endow even (comparatively) smaller models with instruction-following abilities~\cite{sanh2021multitask, wei2021finetuned}. This motivated extensive work on compiling large instruction-tuning datasets, such as the Flan 2021~\cite{chung2022scaling} and Super-NaturalInstructions collections~\cite{wang2022super}, each encompassing over 1600 NLP tasks, and OPT-IML collection with 2000 tasks~\cite{iyer2022opt}.

\paragraph{LLM Prompt Sensitivity} 
However, LLMs are sensitive to how prompts are constructed~\citep{tjuatja_llms_2023, raj_semantic_2023}. In few-shot learning, factors such as the prompt format~\citep{sclar_quantifying_2023,chakraborty_zero-shot_2023}, as well as the choice~\citep{gutierrez2022thinking} and ordering~\citep{lu_fantastically_2022,pezeshkpour_large_2023} of exemplars have a significant impact on task performance. 
In zero-shot settings, \citet{webson_prompt-based_2022} found that models often realize similar performance with misleading or irrelevant prompts as with correct ones. Elsewhere, \citet{sun_evaluating_2023} showed that general domain instruction-tuned LLMs are not robust to variations in instructions --- specifically, they found that models underperform when given novel instructions unseen in training. Our work contributes to this line of research by focusing on the clinical domain.

\paragraph{LLMs for Clinical Tasks}

General domain LLMs encode a surprising amount of clinical and biomedical knowledge allowing them to solve various prediction and information extraction tasks via natural language instructions~\cite{singhal2023large, agrawal2022large,munnangi2024onthefly}. However, smaller models fine-tuned on task-specific data can outperform generalist LLMs in clinical tasks~\cite{lehman2023we}. At the same time, there is a dearth of large high-quality clinical text datasets to train LLMs due to privacy considerations. Researchers have tried to overcome this by exploiting synthetic data generated from biomedical and clinical literature and question answering datasets to train domain-specific models~\cite{toma2023clinical, kweon2023publicly, han2023medalpaca}. However, the resulting  models are often outperformed by general domain variants~\cite{Veen2023AdaptedLL,excoffierGeneralistEmbeddingModels2024} --- our experimental results confirm these observations.

In a contemporaneous study \citet{Chang2024.04.05.24305411} convened a panel of 80 multidisciplinary experts to red team ChatGPT models for the appropriateness of the responses in medical use cases. Experts were asked to write (non-adversarial) prompts for clinically relevant scenarios and the responses were judged by medical doctors  with respect to safety, privacy, hallucinations, and bias. This work is complementary to ours in that it aims to stress test models for the \emph{appropriateness} of their responses to healthcare related prompts whereas we focus on their \emph{sensitivity} to prompt variations.

\section{Conclusions}


This paper presents a large-scale evaluation of instruction-tuned open-source LLMs for clinical classification and information extraction tasks on clinical notes (from EHR). 
We specifically focus on model robustness to natural differences in prompts written by medical professionals. 
We recruited 12 practitioners with different professional and demographic backgrounds, medical specialties, and years of experience to write prompts for 16 clinical tasks spanning binary classification, outcome prediction, and information extraction. 

There are a few main generalizable takeaways relevant to machine learning in healthcare in this work. 
First, the performance LLMs realize on the same clinical task varies substantially across prompts written by different domain experts, and this holds across all models. 
Second, the domain-specific (clinical) models we evaluated perform, in general, worse than their general domain counterparts. Third, prompt variations have concerning implications for fairness — we find that alternative prompts yield different levels of fairness.
Based on these findings, we recommend that practitioners exercise caution when using instruction-tuned LLMs for high stakes clinical tasks which may ultimately impact patient health.
Crucially, clinicians using LLMs should be made aware that subtle, plausible variations in phrasings may yield quite different outputs. Beyond healthcare, this work enriches our understanding of (the lack of) LLM robustness and---we hope---will motivate research into new methods to improve models in this respect.  
\section{Limitations}

Our study reveals that open-source instruction-tuned LLMs are sensitive to instruction phrasings and suggests caution in adopting these models for applications that may impact personal health and well-being. However, this work has several limitations.  First, we acknowledge that our findings may not generalize to larger commercial models but cost and privacy considerations may preclude the deployment of proprietary models for real-world healthcare applications.
Second, we endeavored to recruit a diverse group of medical professionals but our final pool of participants may not be a representative sample of the potential users of these technologies. Moreover, participants were not allowed to see the results of their instructions but in the real world users would have the opportunity to experiment with different prompts and learn how to best use these models. Third, our evaluation protocol for classification tasks may not reflect real world usage --- we induced model predictions from the logit distribution of the first generated token. However, in practice users can only see the final generated outputs and must be able to parse and interpret these in the context of the task at hand.
Finally, our analysis showed that variations in instructions have implications for fairness with respect to race and gender. However, we did not examine the impact of these disparities on intersectional identities which are often affected by compounded biases.

\section*{Acknowledgments}
This work was supported in part by National Science Foundation (NSF) award 1901117, and by the National Insitutes of Health (NIH) award R01LM013772. 

We also thank the reviewers, for their valuable feedback and comments that helped improve this work.

\bibliography{anthology,custom}
\bibliographystyle{acl_natbib}

\clearpage
\appendix
\section{Appendix}

\subsection{Instruction Collection}
\label{apx:anno}

\begin{figure*}
  \centering
  {\includegraphics[width=\textwidth]
  {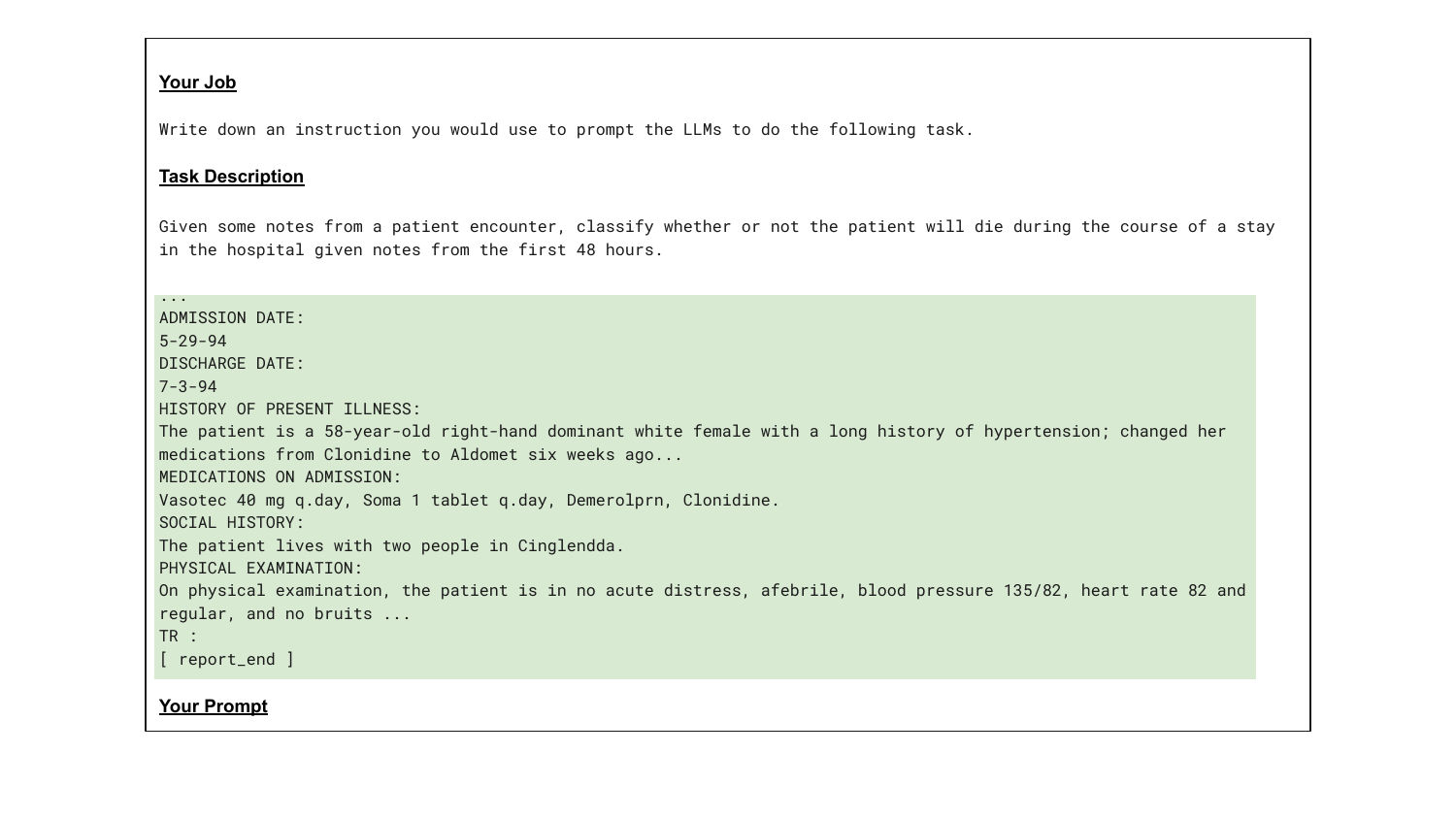}}
  \vspace{-1.5cm}
  \caption{Example of instructions for annotators for a classification task; we provided participants with a description of the tasks including the goal, the expected outputs and a (fictitious) example of a clinical note.}
  \label{fig:anno_ex} 
\end{figure*}

\begin{figure*}
  \centering
  {\includegraphics[width=\textwidth]
  {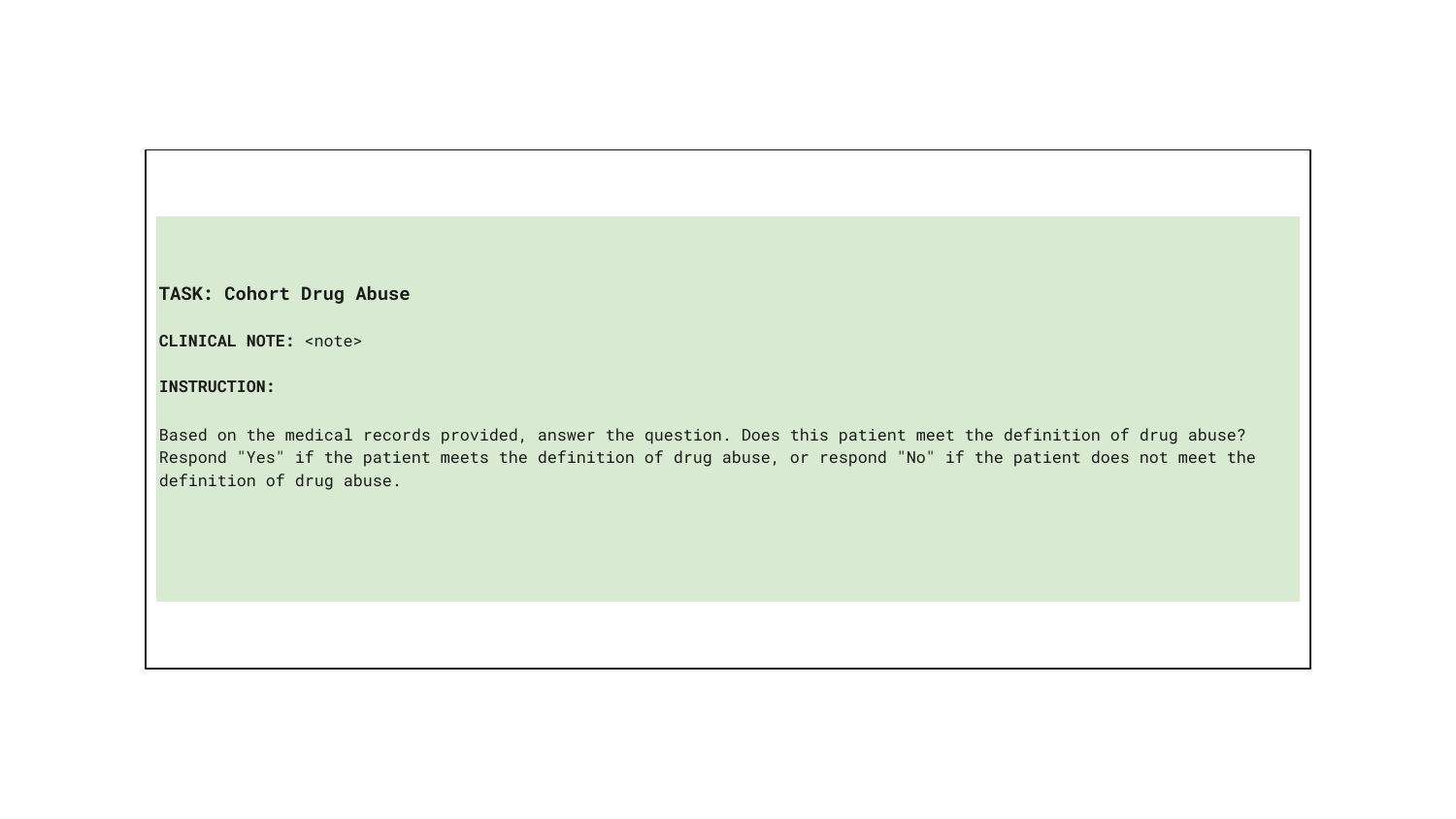}}
  \vspace{-1.5cm}
  \caption{Example of instructions for `Cohort drug abuse' classification task.}
  \label{fig:prompt1} 
\end{figure*}

\begin{figure*}
  \centering
  {\includegraphics[width=\textwidth]
  {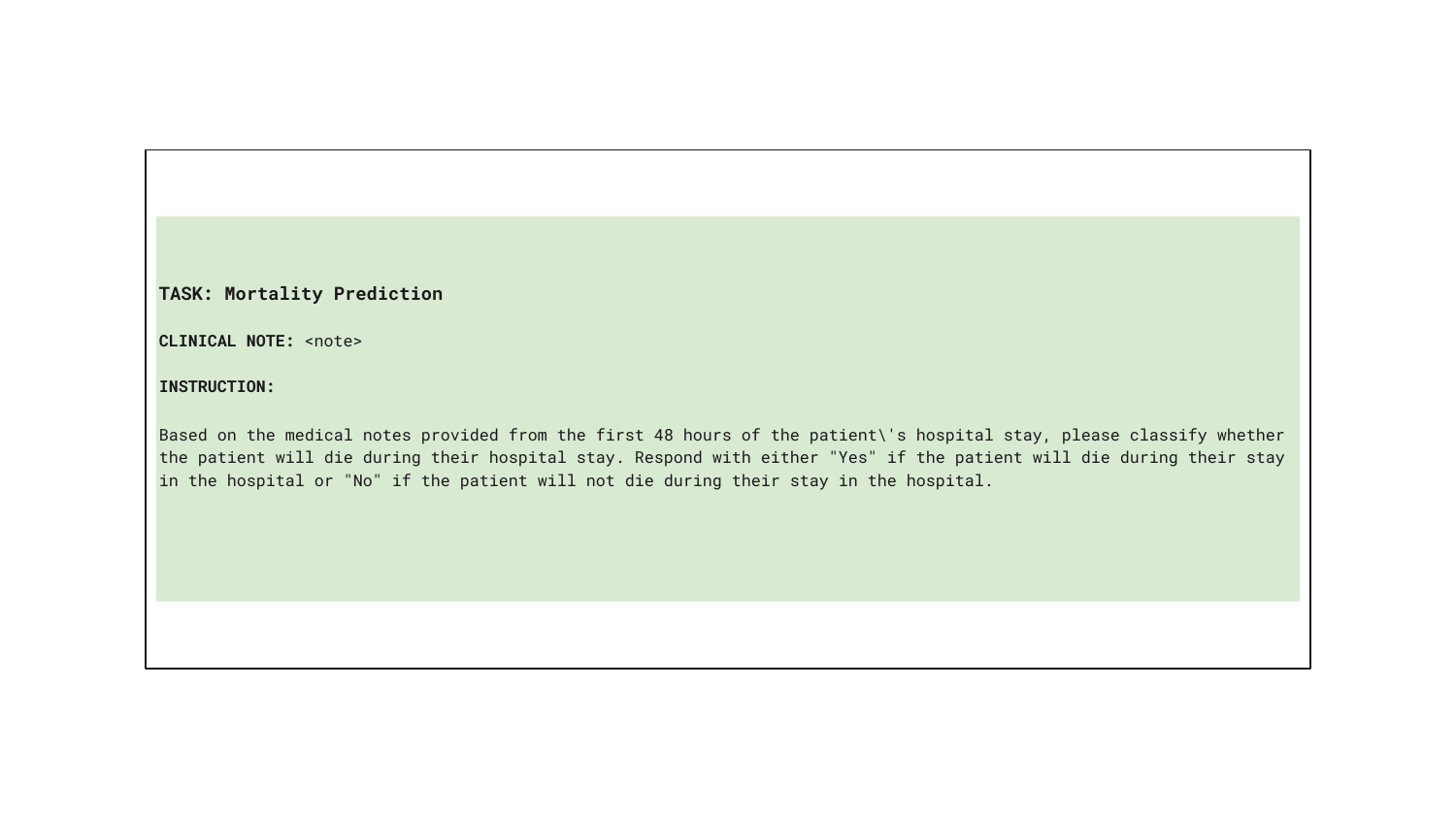}}
  \vspace{-1.5cm}
  \caption{Example of instructions for `Mortality Prediction' classification task.}
  \label{fig:prompt2} 
\end{figure*}

To collect instructions from experts, we provided them with a description of the tasks including the goal, the expected outputs and a (fictitious) example of a clinical note. Figure \ref{fig:anno_ex} is an example of the instructions given for a classification task; and Figures \ref{fig:prompt1} and~\ref{fig:prompt2} show examples of collected instructions. We released the full set of collected instructions along with code.

\subsection{Results}
\label{apx:results}

In this section we present additional results from our experiments. We show detailed results in terms of the mean performance and standard deviation for all the classification and information extraction tasks in tables \ref{tab:main-classification} and \ref{tab:main-extraction}, respectively. 

\begin{figure*}
  \centering 
  \makebox[\textwidth][c]
  {\includegraphics[width=\textwidth]{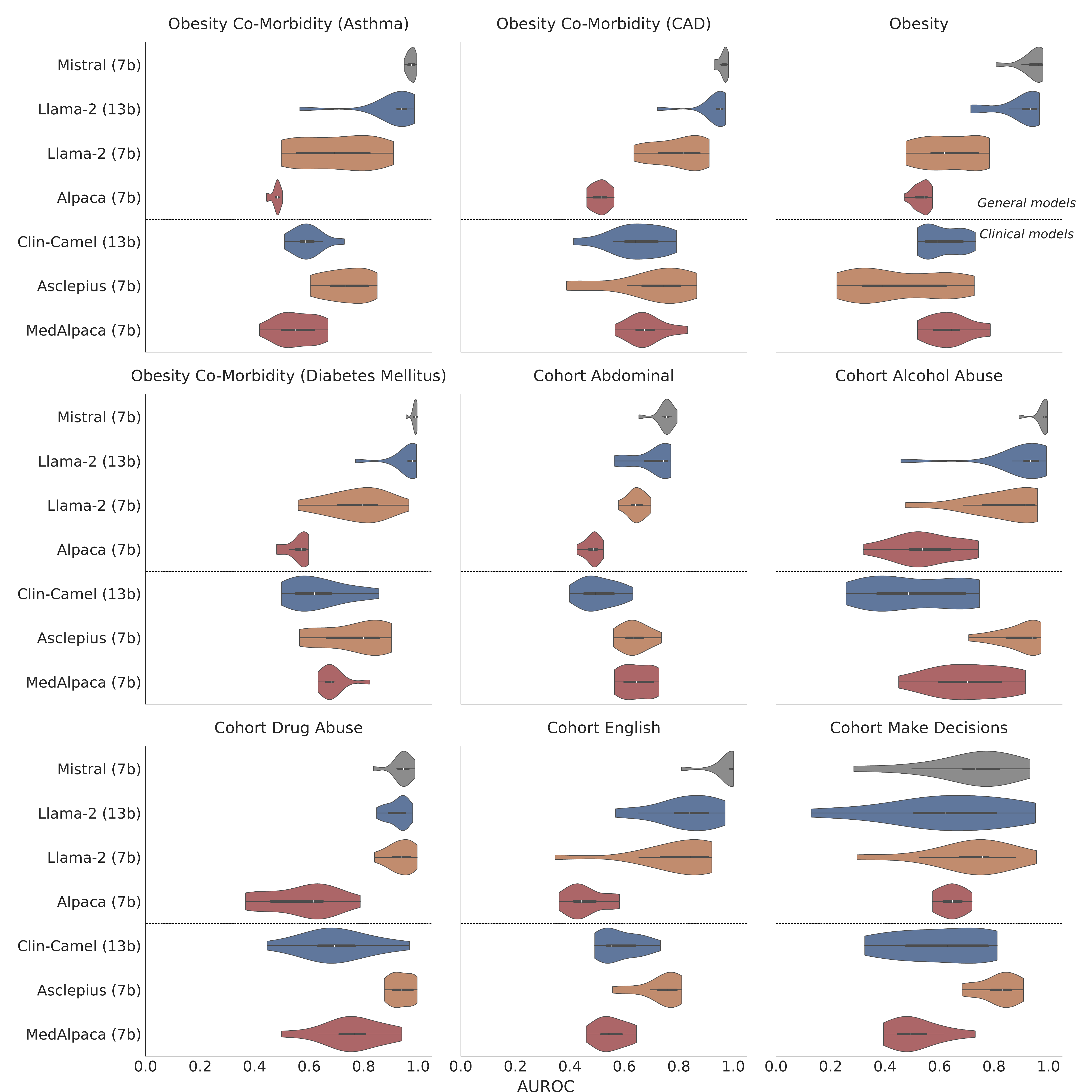}}
  \caption{Variability in performance across prompts for binary classification tasks. Again we observe that different (equivalent) instructions yield wide variances in performance, suggesting an undue sensitivity to phrasings.}
  \label{fig:classification} 
\end{figure*}

\begin{figure*}
\centering
\includegraphics[width=0.96\linewidth]{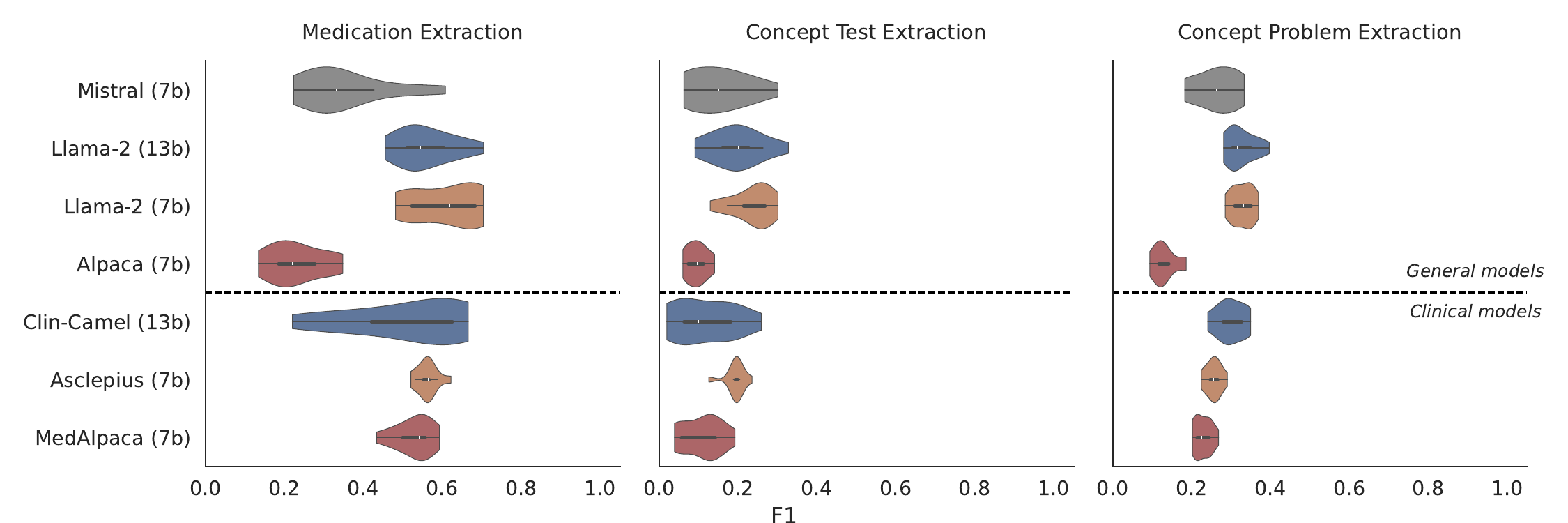} 
  \includegraphics[width=0.64\linewidth]{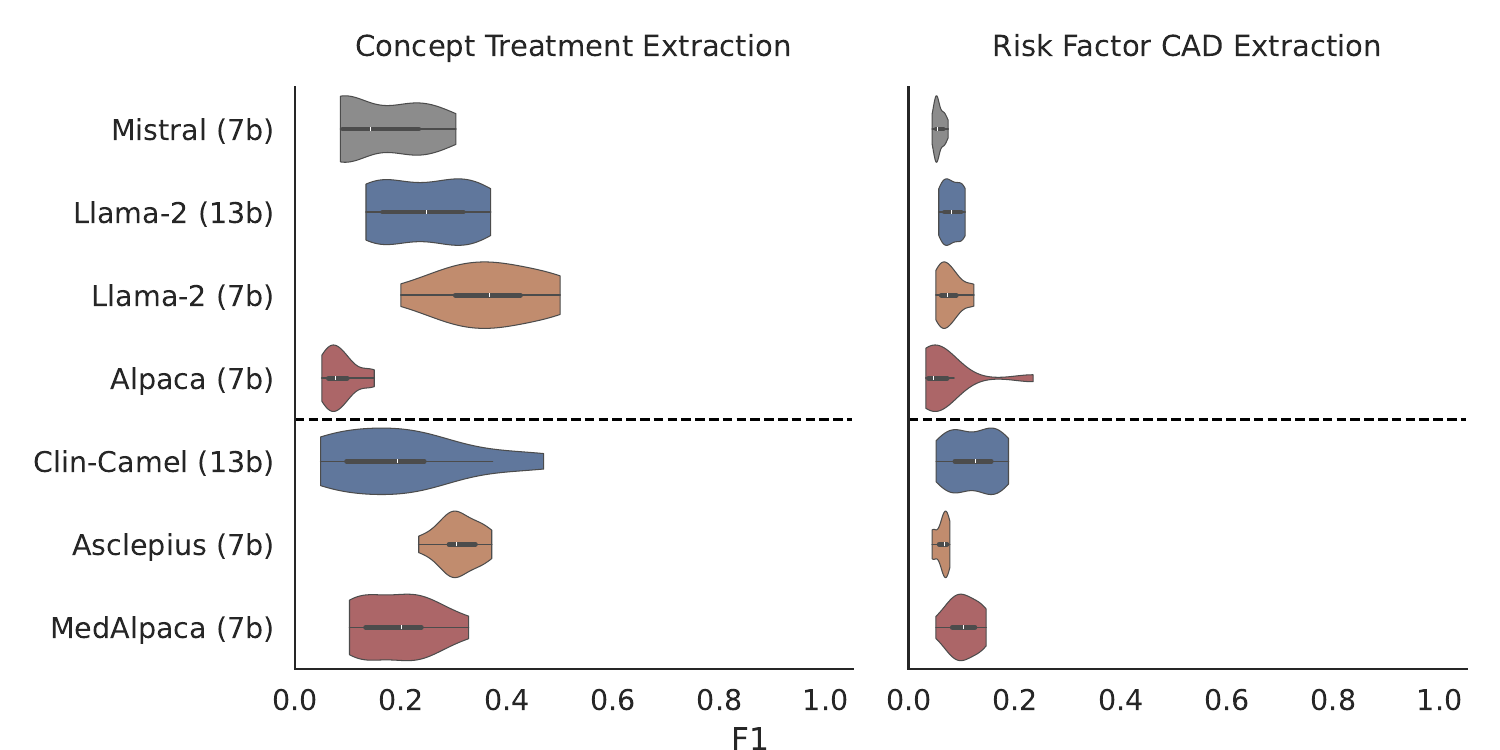}
  \caption {Variability in performance across prompts for the remaining 5 extraction tasks. As mentioned, for most models, different but semantically equivalent prompts yield quite a range of performance.}
  \label{fig:extraction} 
\end{figure*}

\begin{figure*}[th]%
    \centering
    \subfloat{{\includegraphics[width=0.8\textwidth]{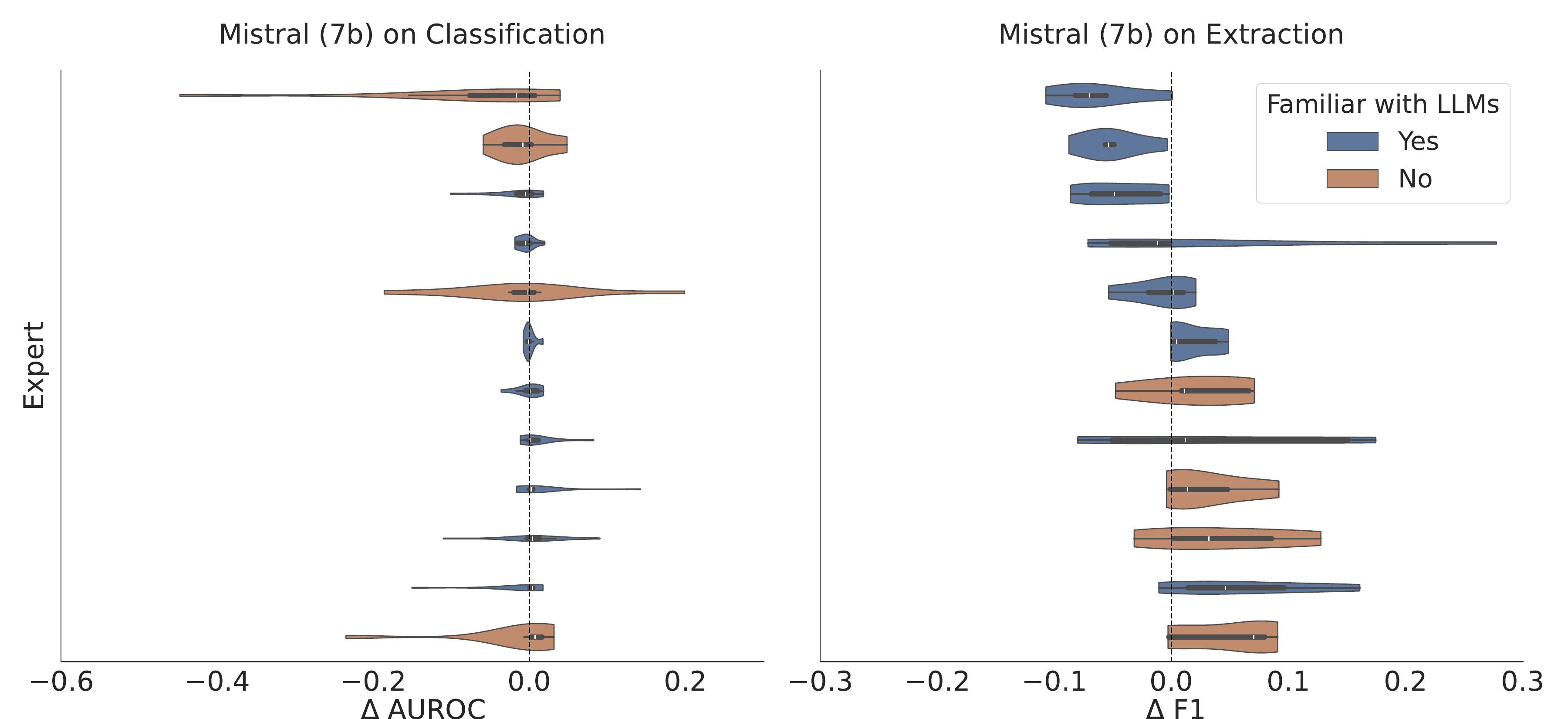} }} \\
    \vspace{1em}
    \subfloat{{\includegraphics[width=0.8\textwidth]{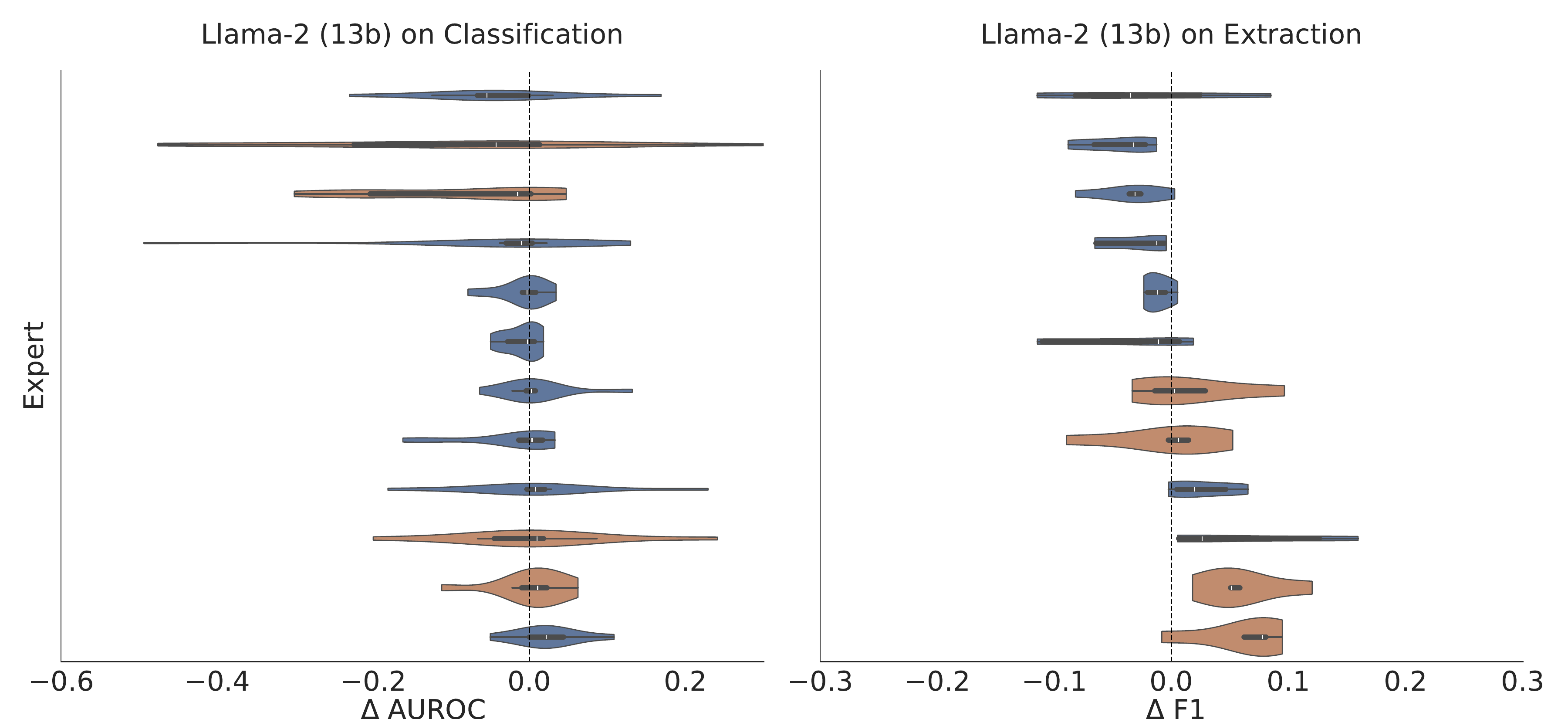} }}%
    \vspace{1em}
    \subfloat{{\includegraphics[width=0.8\textwidth]{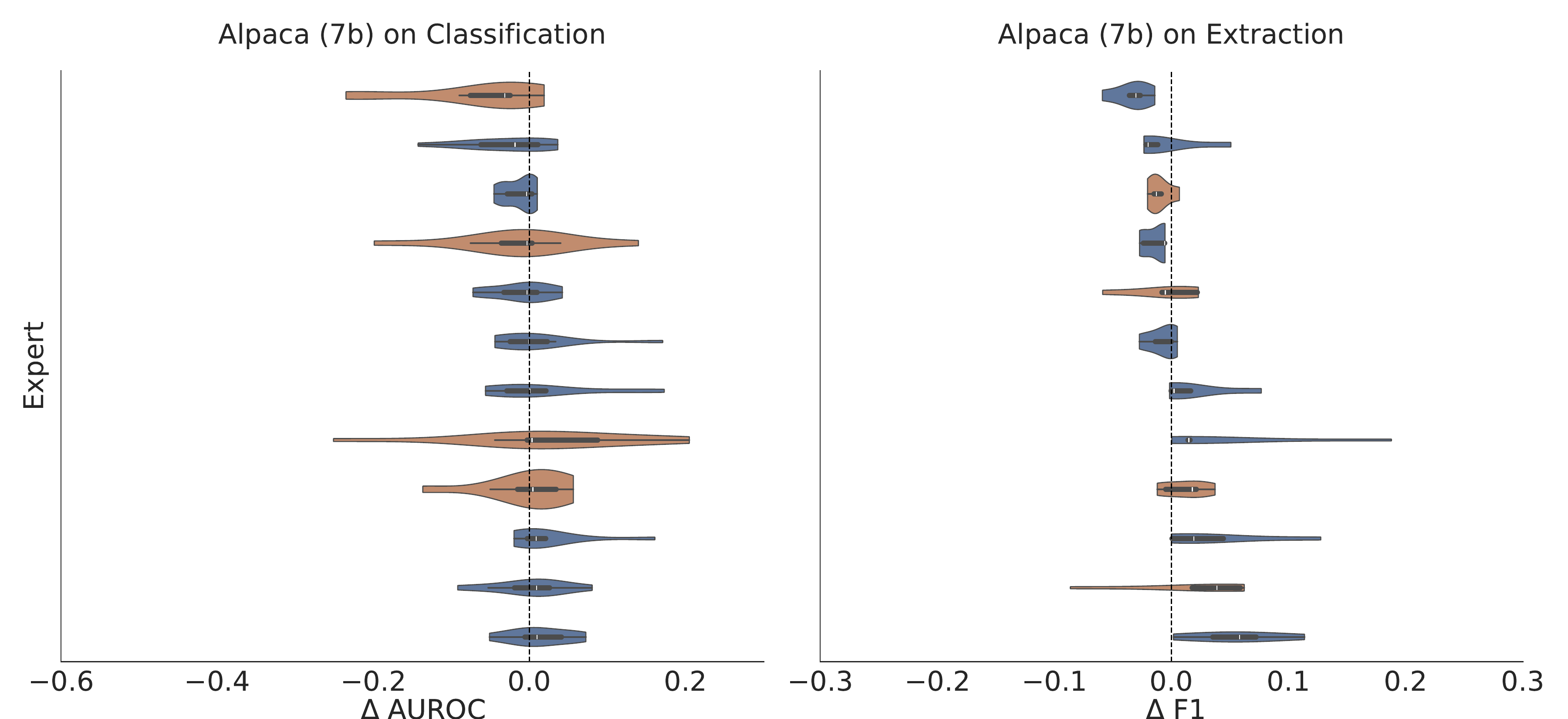} }}%
    \caption{Distribution of performance deltas between each expert’s prompt and the median prompt across all tasks for each \textit{general} model. Each violin plot represents an expert color-coded according to their familiarity with LLM. }
    \label{fig:annotators_analysis_general}%
\end{figure*}

\begin{figure*}[th]%
    \centering
\subfloat{{\includegraphics[width=0.8\textwidth]{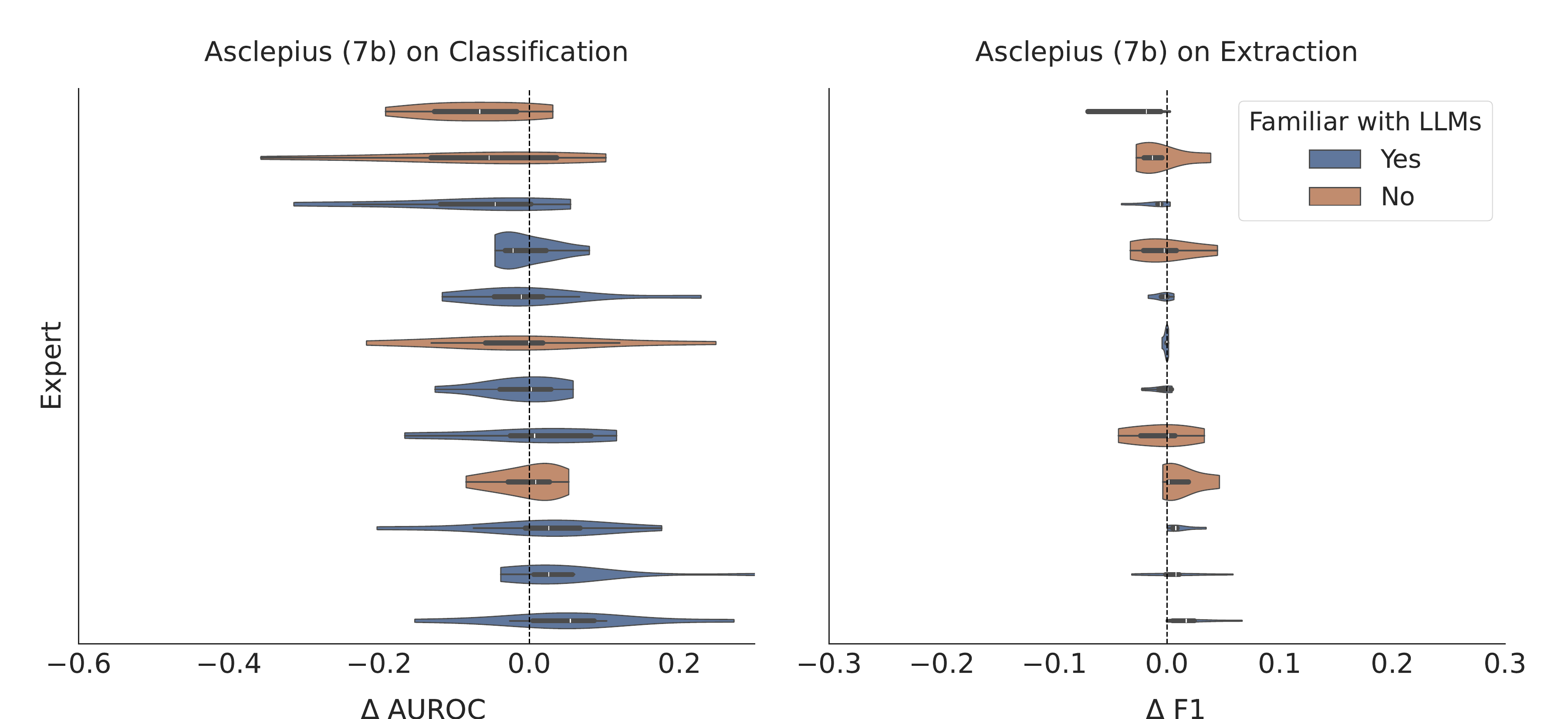} }}%
    \vspace{1em}    \subfloat{{\includegraphics[width=0.8\textwidth]{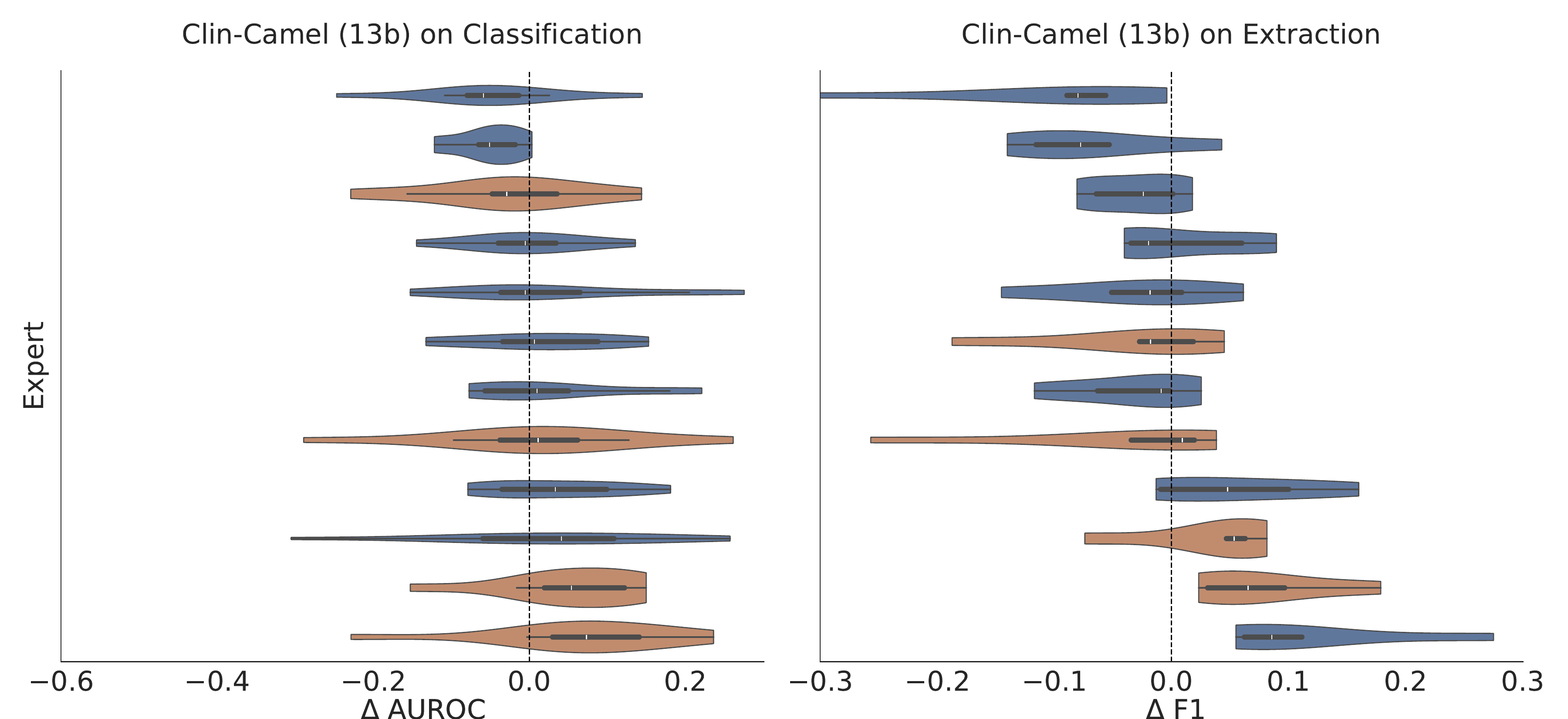} }}%
    \vspace{1em}
\subfloat{{\includegraphics[width=0.8\textwidth]{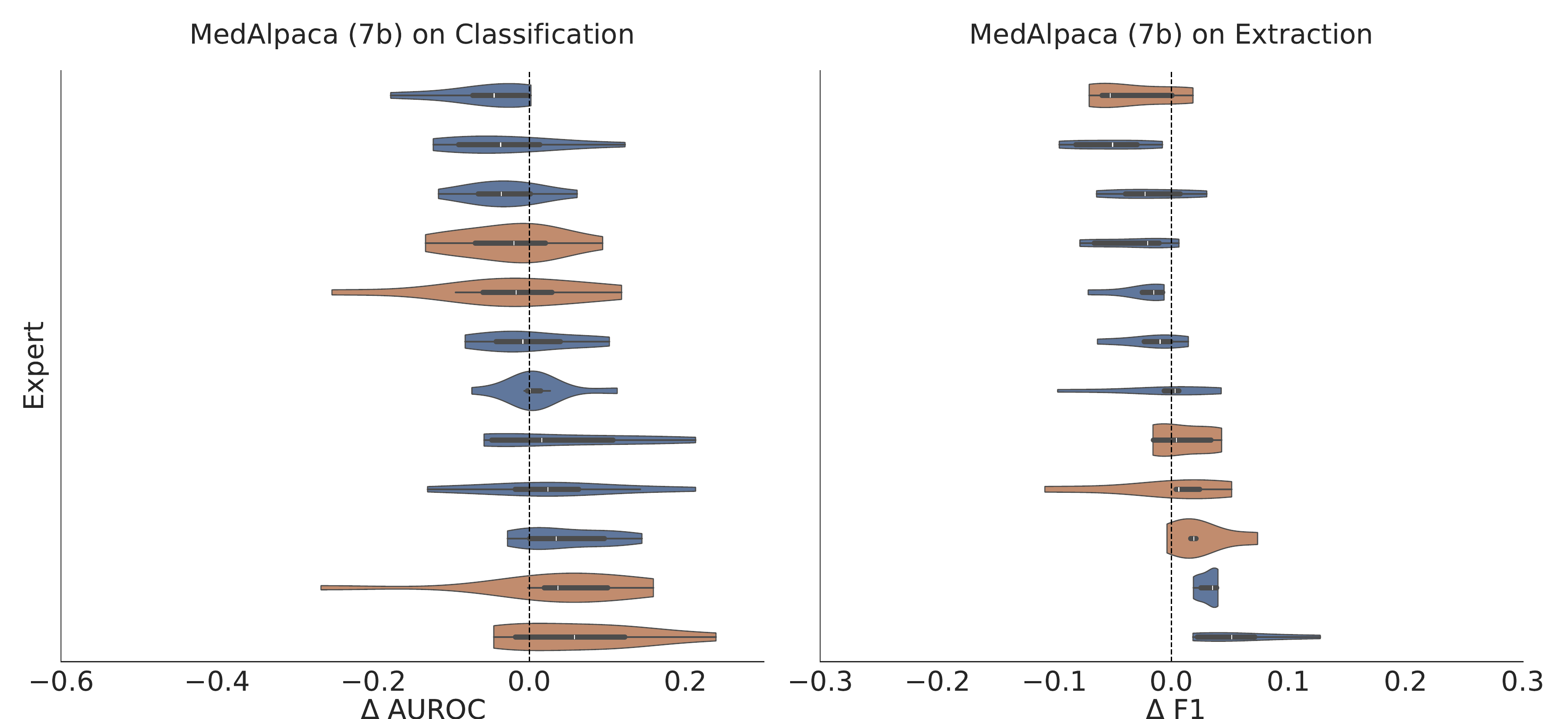}}}%
    \caption{Distribution of performance deltas between each expert’s prompt and the median prompt across all tasks for each \textit{clinical} model. Each violin plot represents an expert color-coded according to their familiarity with LLM.}
    \label{fig:annotators_analysis_clinical}%
\end{figure*}

Figures \ref{fig:classification} and \ref{fig:extraction} plot the variability in performance across classification and extraction tasks, respectively. Figures \ref{fig:annotators_analysis_general} and \ref{fig:annotators_analysis_clinical} plot the deltas in performance between individual expert’s prompts and the median prompt per task, for general domain and clinical models, respectively. 

\begin{figure*}
    \centering
    \subfloat{{\includegraphics[width=7cm]{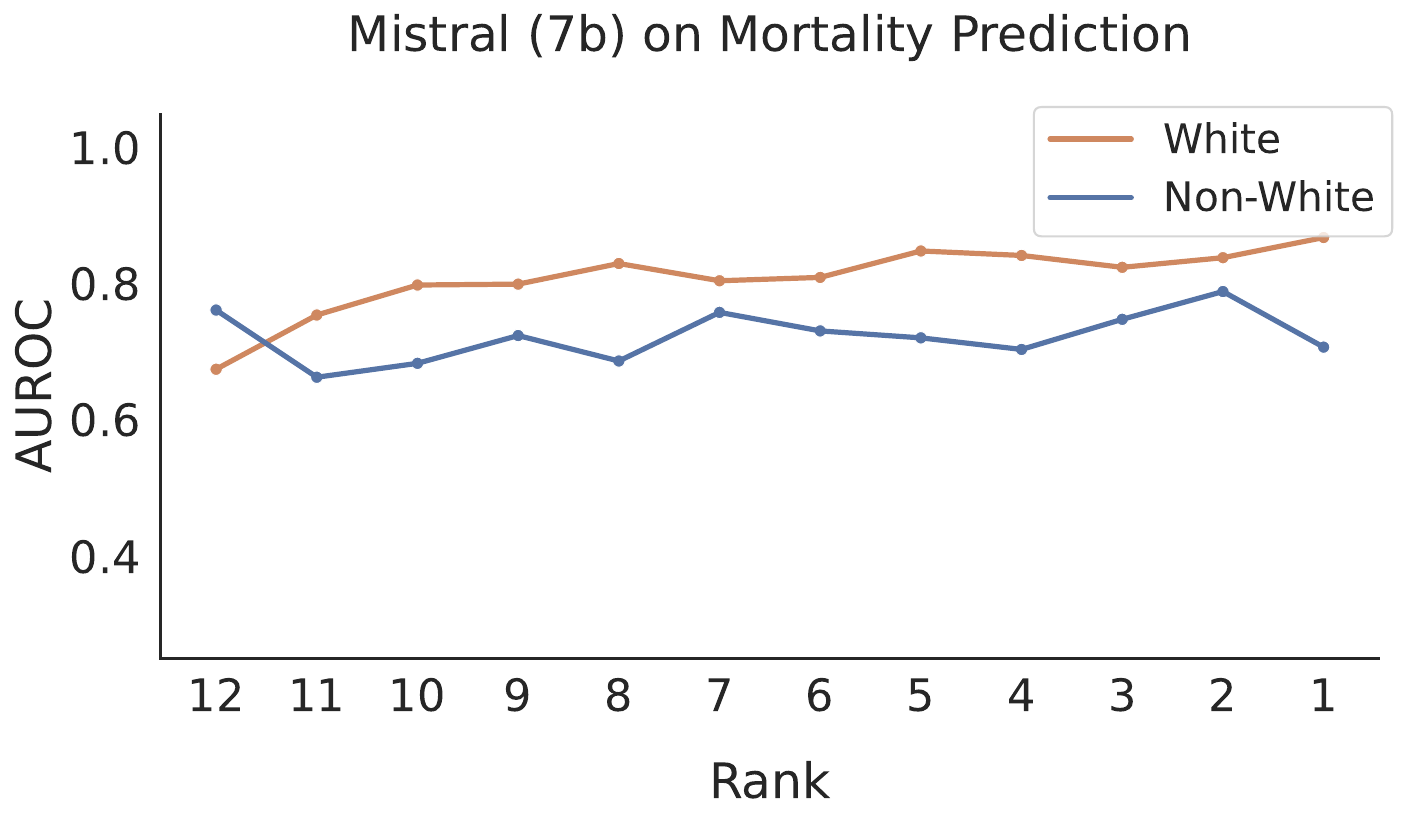} }} \\
    \subfloat{{\includegraphics[width=7cm]{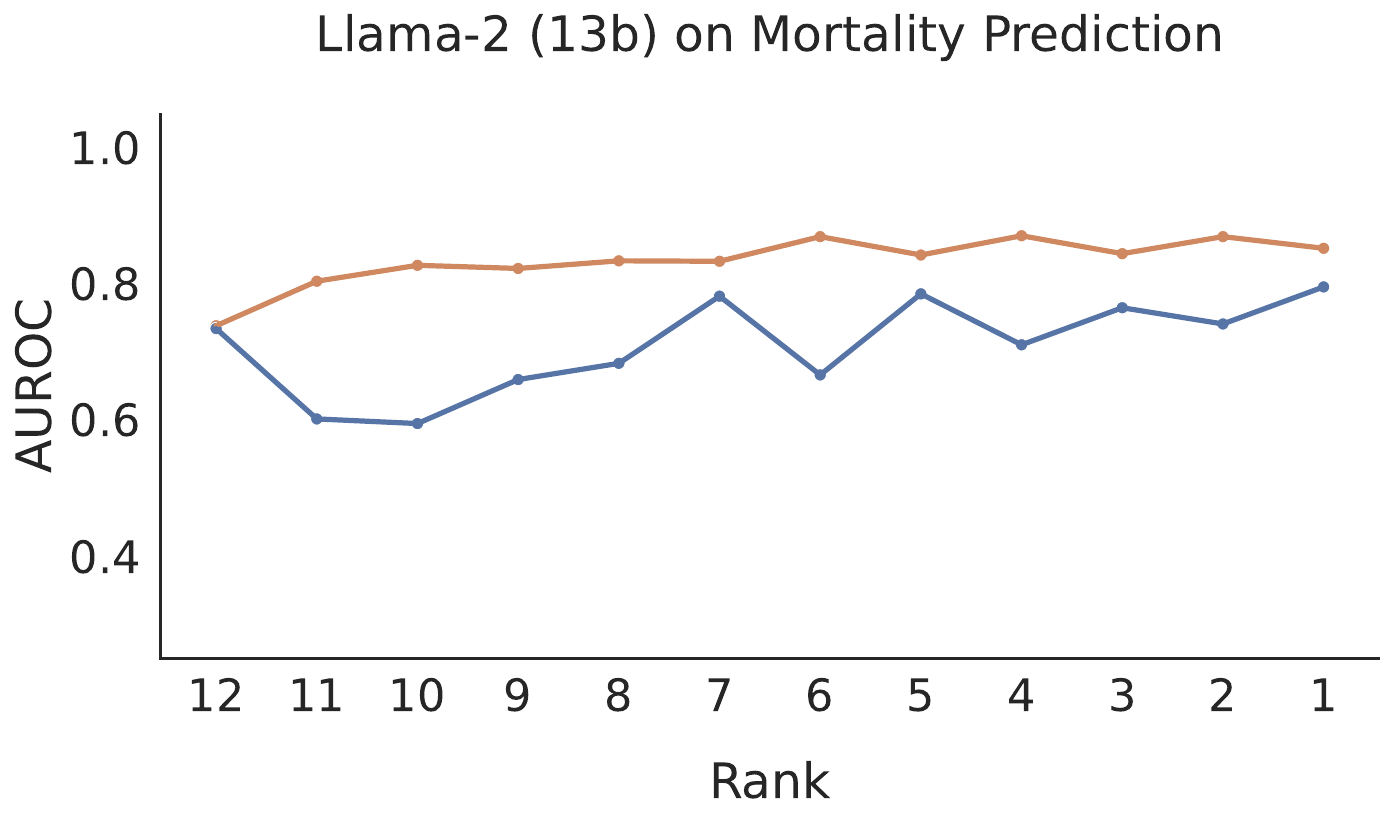} }}%
    \hspace{0em}
    \subfloat{{\includegraphics[width=7cm]{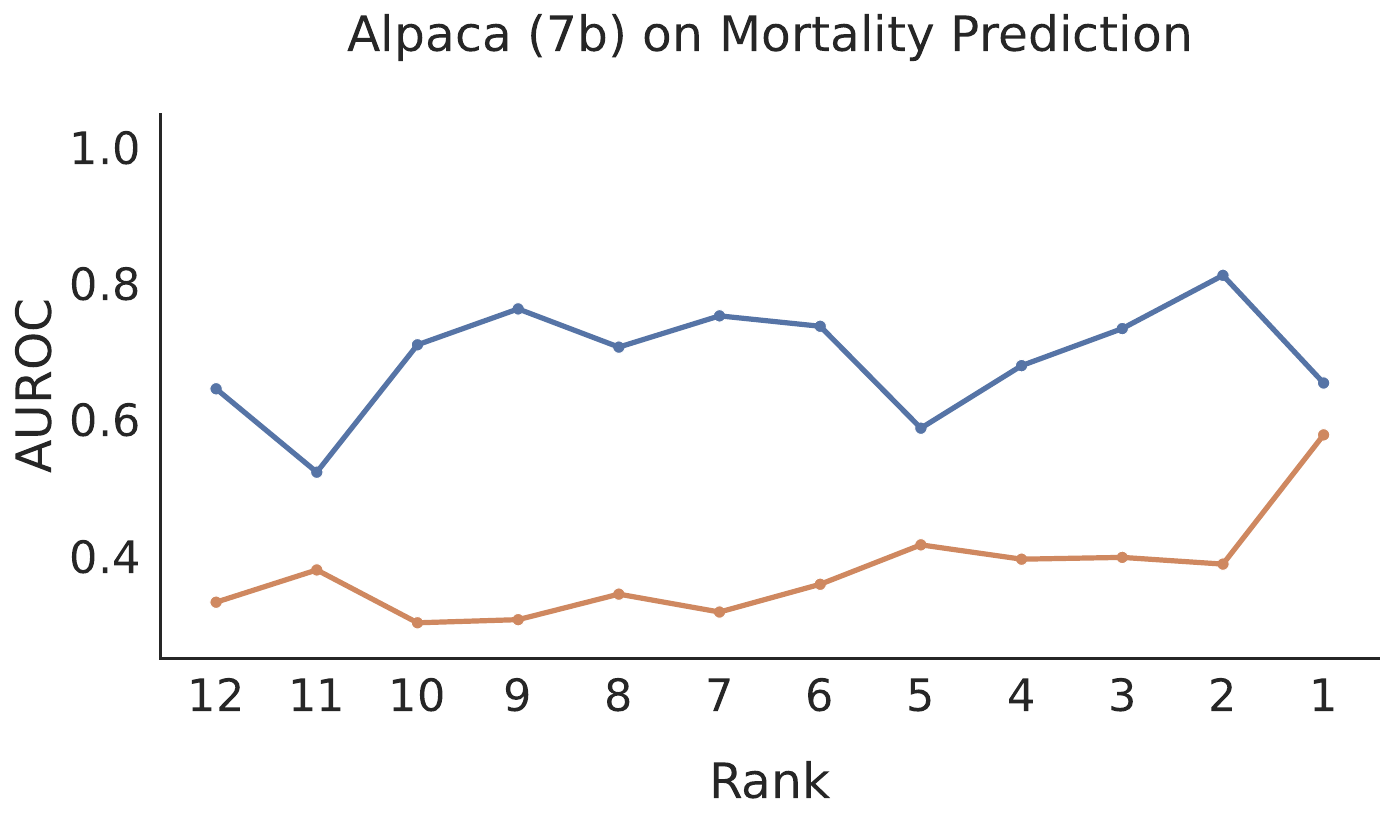} }}%
    \hspace{0em}
    \subfloat{{\includegraphics[width=7cm]{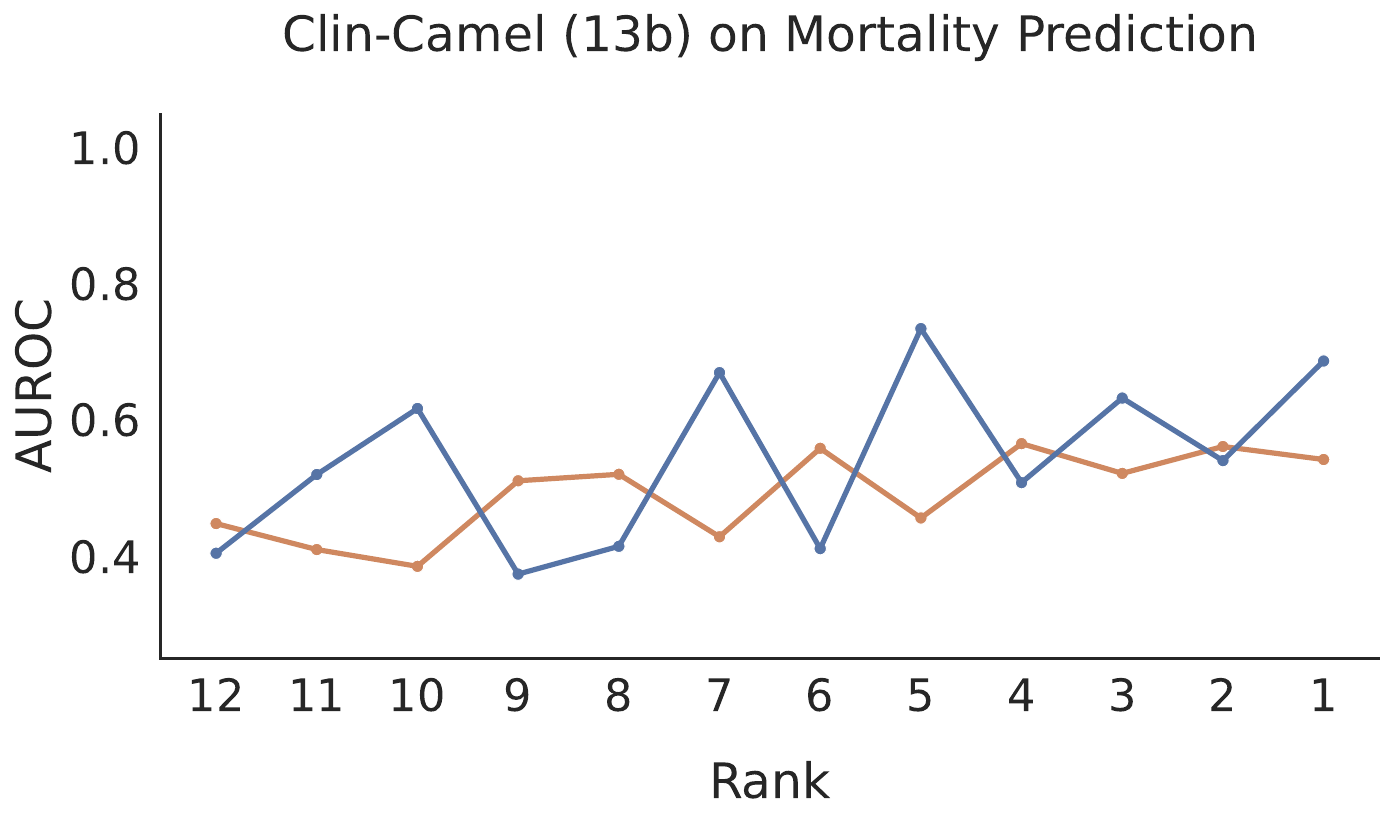} }}%
    \hspace{0em}
    \subfloat{{\includegraphics[width=7cm]{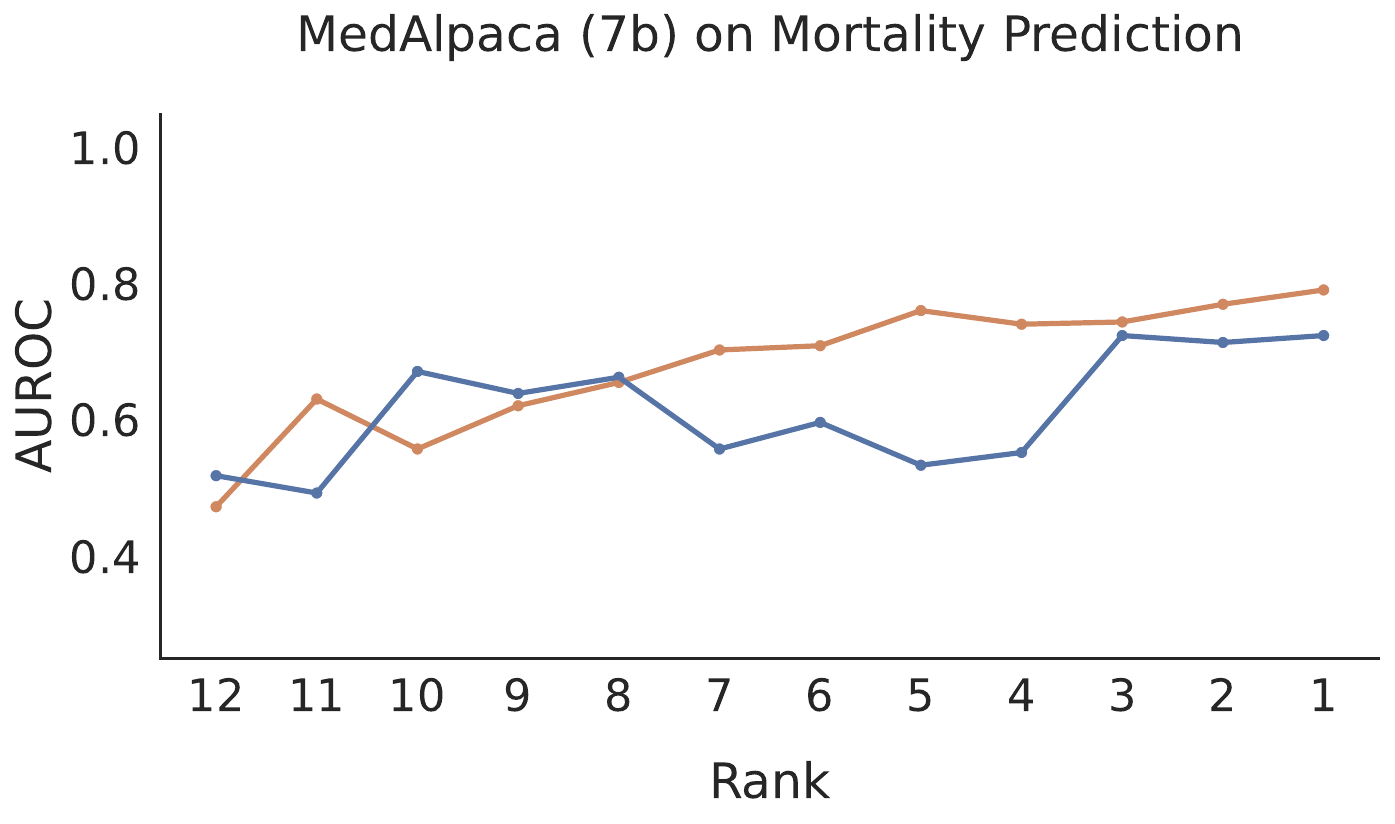} }}%
    \caption{Race subgroup performance on the Mortality Prediction task with a general (left) and clinical model (right). Mistral has no clinical counterpart in our study.}%
    \label{fig:race_analysis_full}%
\end{figure*}

\begin{figure*}[th]%
    \centering
    \subfloat{{\includegraphics[width=7cm]{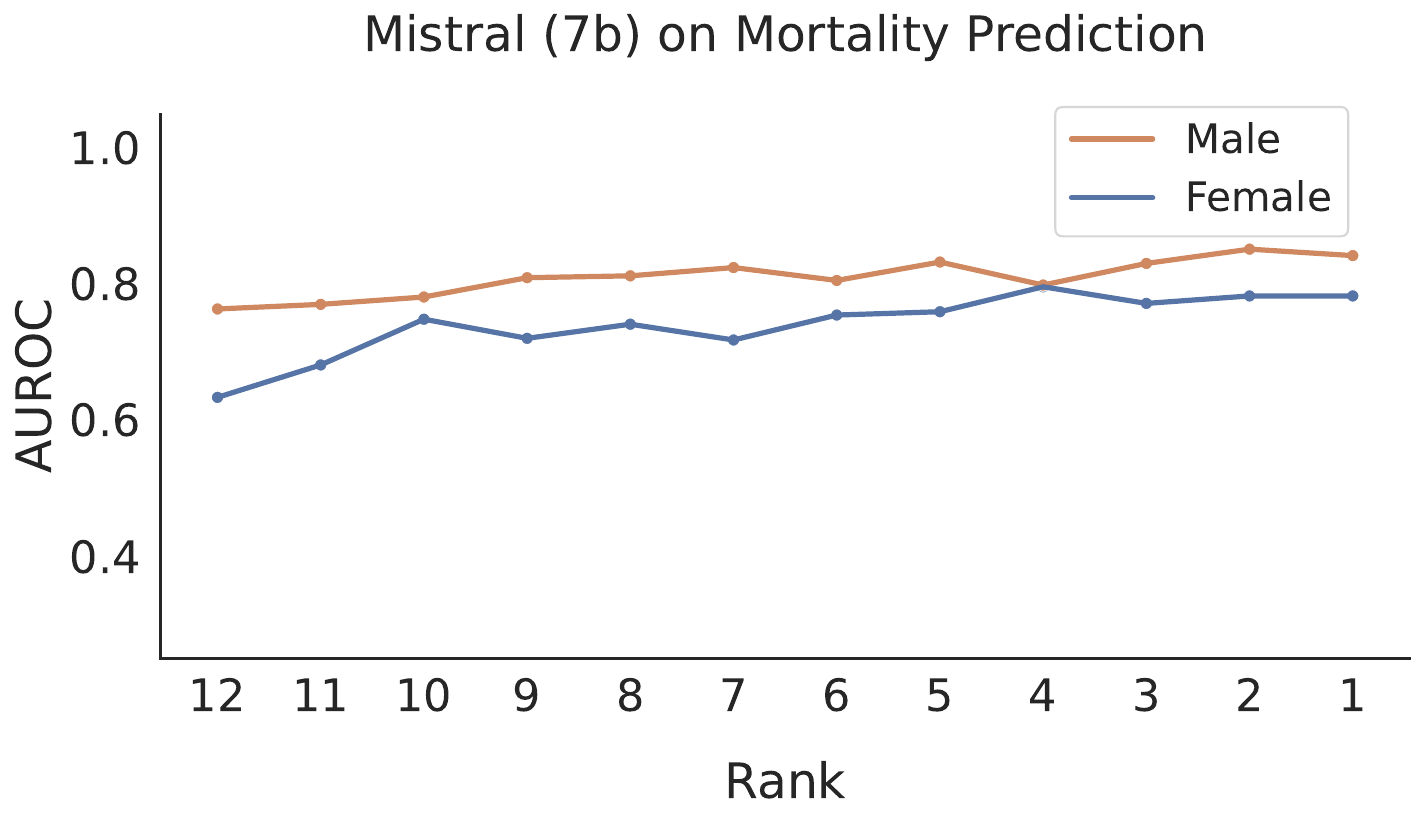} }} \\
    \subfloat{{\includegraphics[width=7cm]{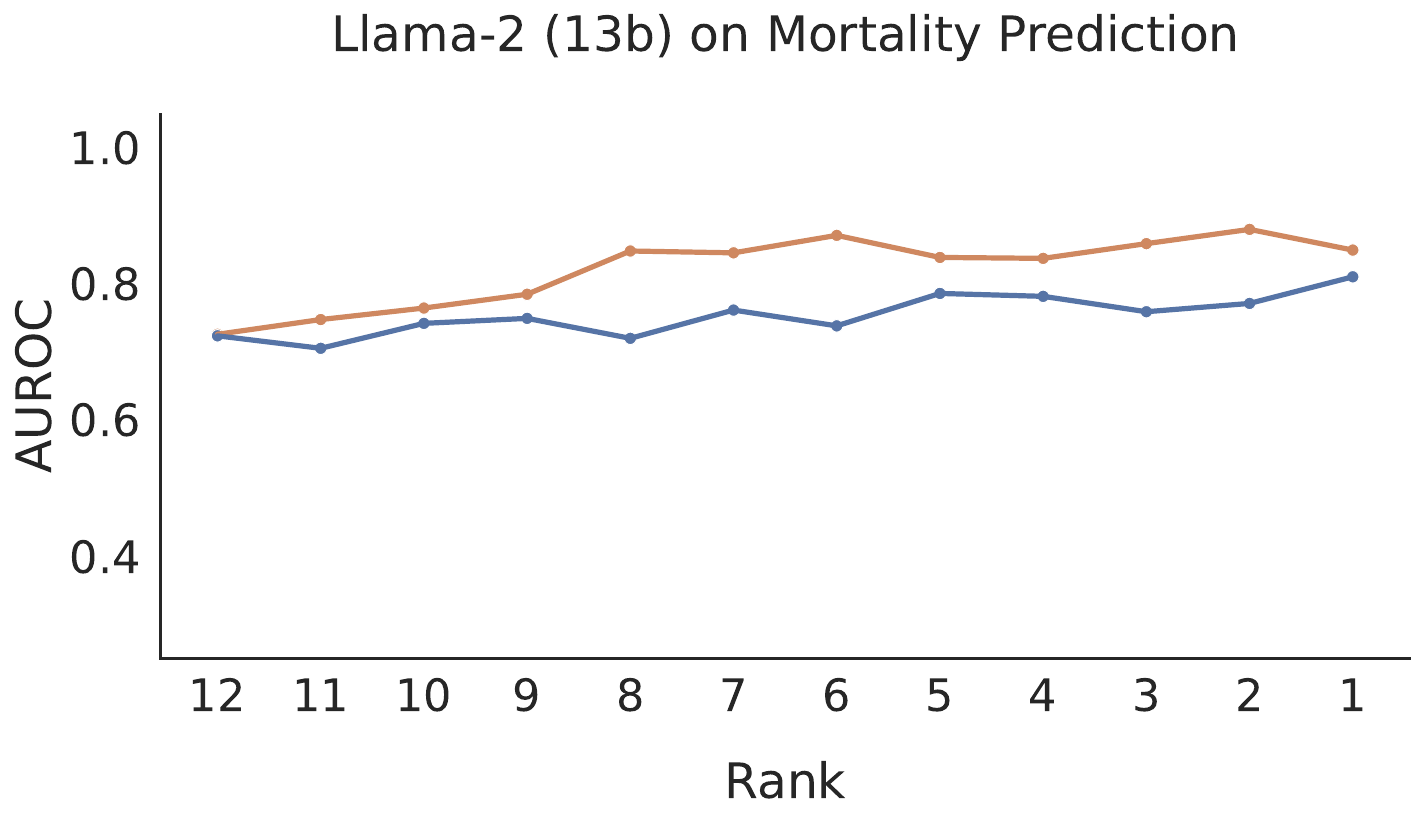} }}%
    \hspace{0em}
    \subfloat{{\includegraphics[width=7cm]{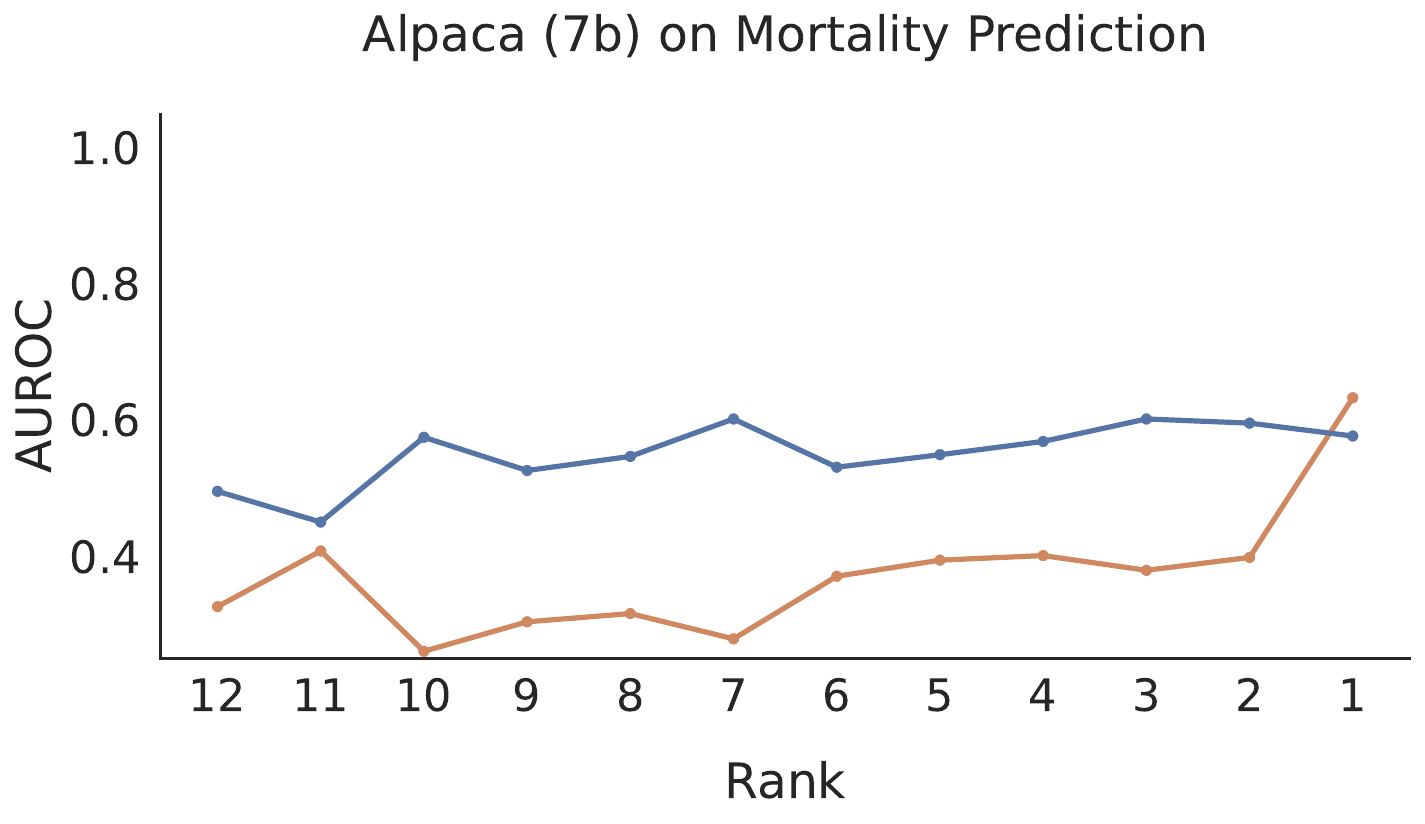} }}%
    \hspace{0em}
    \subfloat{{\includegraphics[width=7cm]{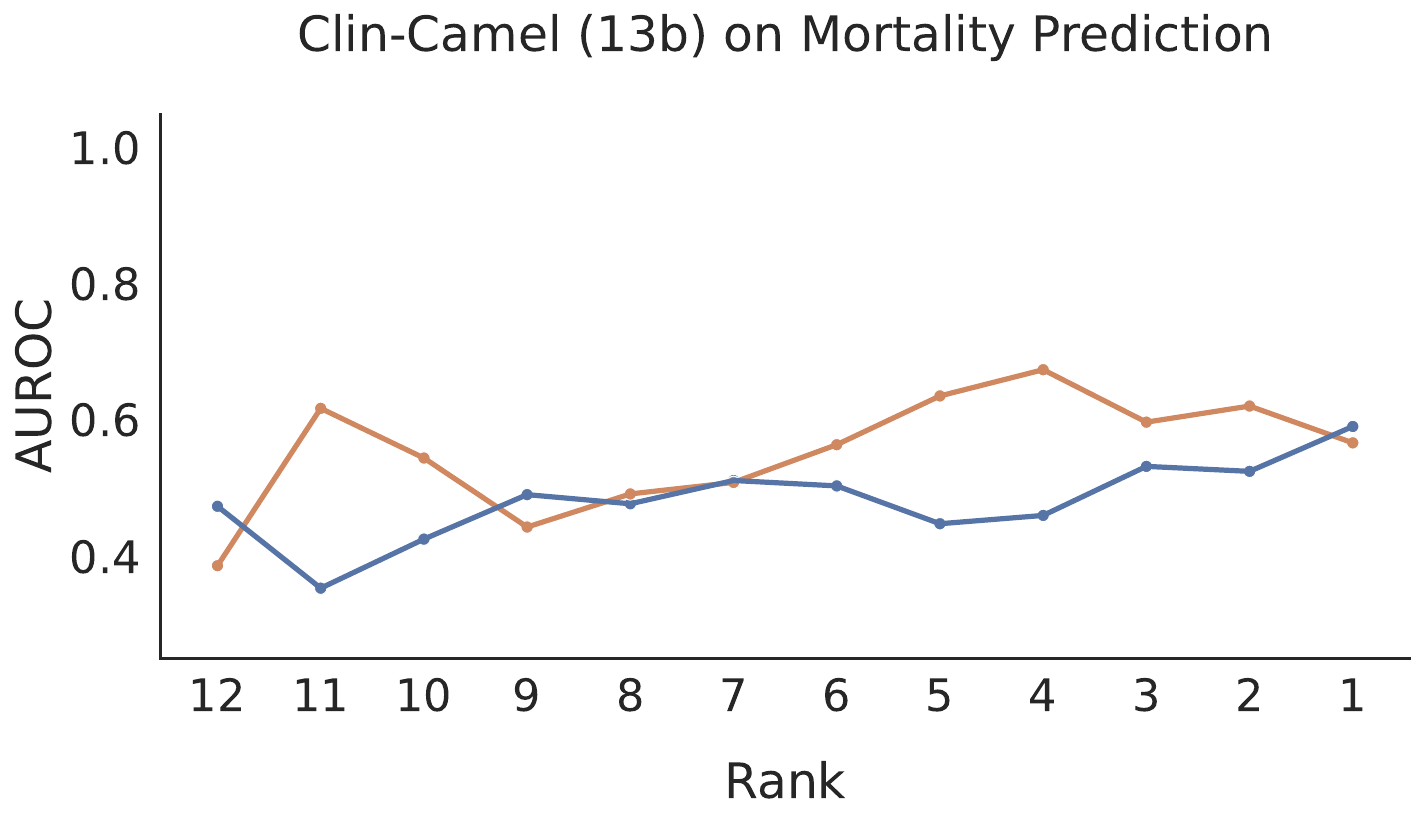} }}%
    \hspace{0em}
    \subfloat{{\includegraphics[width=7cm]{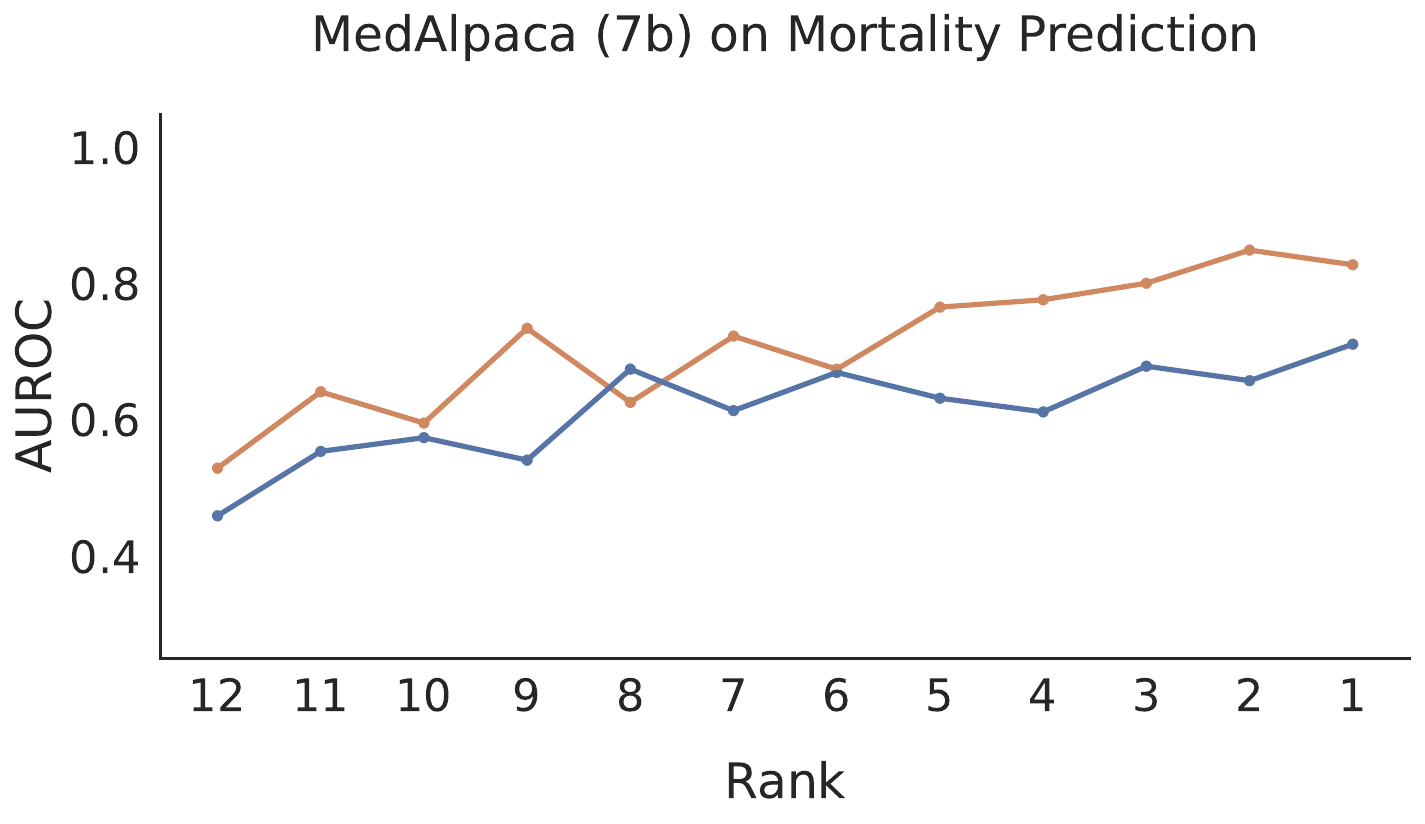} }}%
    \caption{Sex subgroup performance on the Mortality Prediction task with a general (left) and clinical model (right). Mistral has no clinical counterpart in our study.}
    \label{fig:sex_analysis_full}%
\end{figure*}

Figure \ref{fig:race_analysis_full} show race subgroup performance for the Mortality Prediction task for all the models, and Figure \ref{fig:sex_analysis_full} shows a similar analysis for sex. 

\begin{figure*}
  \centering
  {\includegraphics[width=0.7\textwidth,trim={0 0 0 0.66cm},clip]
  {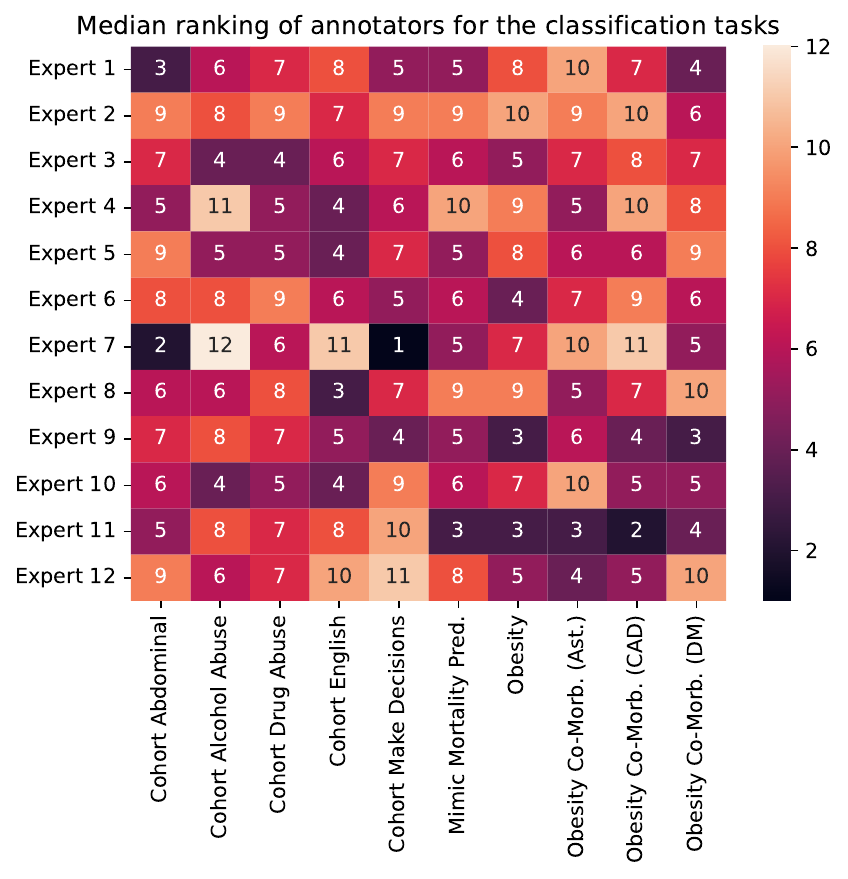}}
  \caption{Median ranking of prompts written by experts for classification tasks across models.}
  \label{fig:consistency_cls} 
\end{figure*}

\begin{figure*}
  \centering
  {\includegraphics[width=0.7\textwidth,trim={0 0 0 0.66cm},clip]
  {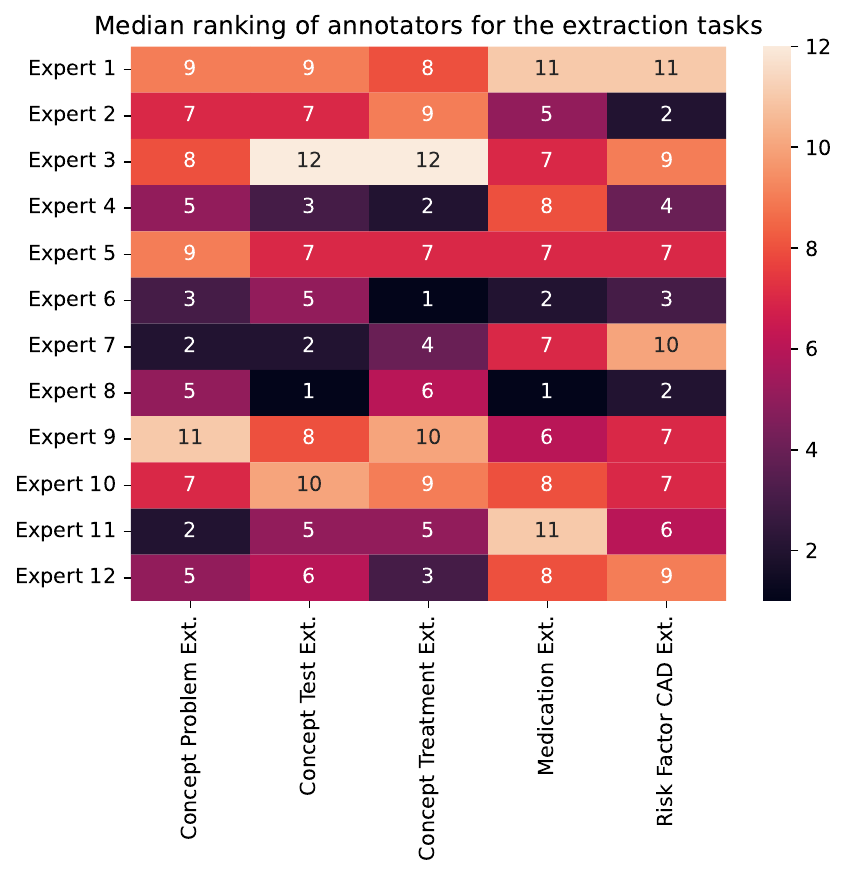}}
  \caption{Median ranking of prompts written by experts for extraction tasks across models.}
  \label{fig:consistency_ext} 
\end{figure*}

Our overall results show that, in general, different prompt phrasings yield different performance. Are there prompts that are consistently effective across models? To investigate this, we ranked each prompt with respect to the performance and calculated the median across models. Figures \ref{fig:consistency_cls} and \ref{fig:consistency_ext} depict the median performance ranking (among all 12 prompts) achieved by the instructions written by each expert. For classification tasks such as Cohort Abdominal and Cohort Make Decisions, Expert 7 wrote prompts that are consistently among the best performing ones for most models, which is also the case for the prompts written by Expert 11 across five classification  tasks. On the other hand, prompts from Expert 2 were consistently among the lower performing ones. A similar pattern can be seen in the extraction tasks, where Experts 6 and 8 wrote some of the best-performing prompts for most of these tasks. This suggests that, to an extent, the performance of prompts is consistent even when tested on different models.

\begin{table*}
    \centering
    \scriptsize
    \begin{tabular}{l c c c c c c c} 
    \toprule
     \textbf{Model / } & \textsc{Mistral} & \textsc{Llama 2 } & \textsc{Llama 2} & \textsc{Alpaca}  & \textsc{Clinical } &  \textsc{Asclepius} & \textsc{MedAlpaca} \\

     \textbf{Dataset} & \textsc{IT 0.2 (7b)} & \textsc{Chat (13b)} & \textsc{Chat (7b)} & \textsc{(7b)} & \textsc{Camel (13b)} & \textsc{(7b)}  & \textsc{(7b)} \\
         \midrule
         Obesity Co- & 0.974 & 0.908 & 0.696 & 0.479  & 0.594 & 0.732 & 0.557 \\
         Morbidity (Asthma) & $\pm(0.014)$  & $\pm(0.111)$ & $\pm(0.145)$ & $\pm(0.017)$  & $\pm(0.059)$ & $\pm(0.086)$ & $\pm(0.078)$ \\
         \cmidrule{2-8}
          Cohort Alcohol & 0.980 &  0.898 &  0.836 & 0.549 & 0.517 & 0.894 & 0.715 \\
         Abuse & $\pm(0.028)$ & $\pm(0.142)$ & $\pm(0.148)$ &  $\pm(0.126)$ & $\pm(0.177)$ & $\pm(0.084)$ & $\pm(0.146)$ \\
         \cmidrule{2-8}
         Obesity Co- & 0.963 & 0.933 & 0.796 & 0.512  & 0.649 & 0.702 & 0.679 \\
         Morbidity CAD &  $\pm(0.017)$  & $\pm(0.067)$ & $\pm(0.096)$ & $\pm(0.033)$  & $\pm(0.107)$ & $\pm(0.154)$ & $\pm(0.071)$ \\
         \cmidrule{2-8}
         Cohort Drug & 0.941 &  0.923 & 0.934  & 0.570  & 0.698 & 0.938 &  0.756 \\
         Abuse & $\pm(0.039)$  & $\pm(0.04)$ & $\pm(0.048)$ & $\pm(0.132)$ & $\pm(0.138)$ & $\pm(0.042)$ & $\pm(0.119)$ \\
         \cmidrule{2-8}
         Cohort English & 0.974  & 0.824  & 0.790 & 0.460 & 0.586 & 0.737 & 0.552 \\
         & $\pm(0.055)$  & $\pm(0.123)$ & $\pm(0.165)$ & $\pm(0.071)$  & $\pm(0.076)$ & $\pm(0.078)$ & $\pm(0.058)$  \\
         \cmidrule{2-8} 
         Cohort Make & 0.709  & 0.623 &  0.710 & 0.644 & 0.597 & 0.817 & 0.513 \\
         Decision & $\pm(0.178)$ & $\pm(0.238)$ &$\pm(0.171)$ & $\pm(0.047)$ & $\pm(0.174)$ & $\pm(0.074)$ & 
         $\pm(0.098)$ \\
         \cmidrule{2-8}
         Cohort & 0.750  & 0.707 & 0.644& 0.483  & 0.506 & 0.637 & 0.648 \\
         Abdominal & $\pm(0.034)$ & $\pm(0.076)$ & $\pm(0.034)$ & $\pm(0.029)$  & $\pm(0.069)$ & $\pm(0.052)$& $\pm(0.059)$ \\
         \cmidrule{2-8}
         Obesity Co- & 0.987 & 0.958 & 0.775  & 0.560 & 0.637 & 0.762 & 0.686 \\
         Morbidity (Diabetes)& $\pm(0.011)$ & $\pm(0.063)$ & $\pm(0.114)$ & $\pm(0.041)$ & $\pm(0.109)$ & $\pm(0.124)$ & $\pm(0.05)$ \\
         \cmidrule{2-8}
         Obesity & 0.943 & 0.9 & 0.639 & 0.534  & 0.612 & 0.453 & 0.64 \\
         Classification & $\pm(0.05)$  & $\pm(0.087)$ & $\pm(0.113)$ & $\pm(0.03)$  & $\pm(0.074)$ & $\pm(0.177)$ & $\pm(0.084)$ \\
         \cmidrule{2-8}
         Mortality & 0.777  & 0.794 & 0.742 & 0.466 & 0.506 & 0.757 & 0.658 \\
         Prediction & $\pm(0.034)$  & $\pm(0.036)$ & $\pm(0.083)$ & $\pm(0.051)$ & $\pm(0.052)$ & 
         $\pm(0.037)$ & $\pm(0.08)$ \\
         \bottomrule
    \end{tabular}
    \caption{Mean and Standard Deviation for instructions on classification tasks across all models and all tasks}
    \label{tab:main-classification}
\end{table*}

\begin{table*}
    \centering
    \scriptsize
    \begin{tabular}{l c c c c c c c}
    \toprule
     \textbf{Model / } & \textsc{Mistral} & \textsc{Llama 2 } & \textsc{Llama 2} & \textsc{Alpaca} & \textsc{Clinical} & \textsc{Asclepius} & \textsc{MedAlpaca} \\

     \textbf{Dataset} & \textsc{IT 0.2 (7b)} & \textsc{Chat (13b)} & \textsc{Chat (7b)} & \textsc{(7b)} & \textsc{Camel (13b)} & \textsc{(7b)} & \textsc{(7b)} \\
        \midrule
         Medication & 0.351 & 0.559 & 0.608 & 0.231 & 0.509 & 0.562 & 0.529 \\
         Extraction & $\pm(0.111)$ & $\pm(0.072)$ & $\pm(0.084)$ &  $\pm(0.069)$ & $\pm(0.15)$ & 
         $\pm(0.027)$ & $\pm(0.047)$ \\
         \cmidrule{2-8}
         Concept Problem & 0.265 & 0.325 & 0.329 & 0.131 & 0.3 & 0.256 & 0.229 \\
         Extraction & $\pm(0.051)$ & $\pm(0.035)$ & $\pm(0.027)$ & $\pm(0.029)$ & $\pm(0.035)$ & $\pm(0.019)$ & $\pm(0.021)$ \\
         \cmidrule{2-8}
         Concept Test & 0.154 & 0.197 & 0.236 & 0.097 & 0.117 & 0.194 & 0.109 \\
         Extraction & $\pm(0.076)$ & $\pm(0.066)$ & $\pm(0.05)$ & $\pm(0.025)$ & $\pm(0.078)$ & 
         $\pm(0.025)$& $\pm(0.049)$ \\
         \cmidrule{2-8}
         Concept Treatment & 0.165 & 0.244 &  0.367 & 0.086 & 0.198 & 0.308 & 0.193 \\
         Extraction & $\pm(0.084)$ & $\pm(0.086)$ & $\pm(0.093)$ & $\pm(0.031)$ & $\pm(0.129)$ & 
         $\pm(0.039)$ & $\pm(0.072)$ \\
         \cmidrule{2-8}
         Drug & 0.394 & 0.373 & 0.495 & 0.192 & 0.372 & 0.432 & 0.429 \\
         Extraction & $\pm(0.101)$ & $\pm(0.047)$ & $\pm(0.072)$ & $\pm(0.074)$ & $\pm(0.128)$ & 
         $\pm(0.042)$ & $\pm(0.086)$ \\
         \cmidrule{2-8}
         Risk Factor CAD & 0.057 & 0.081 & 0.079 & 0.067 & 0.122 & 0.063 & 0.103 \\
         Extraction & $\pm(0.009)$ & $\pm(0.018)$ & $\pm(0.024)$ & $\pm(0.056)$ & $\pm(0.046)$ & $\pm(0.012)$ & $\pm(0.029)$ \\
         \bottomrule
    \end{tabular}
    \caption{Mean and Standard Deviation for instructions on extraction tasks across all models and all tasks}
    \label{tab:main-extraction}
\end{table*}

\end{document}